\documentclass[preprint,12pt]{elsarticle}

\usepackage{amsmath}
\usepackage{amssymb}
\usepackage{amsthm}
\usepackage{mathtools}
\usepackage{paralist}
\usepackage[misc]{ifsym}
\usepackage{graphicx}
\usepackage[colorlinks=true]{hyperref}
\usepackage[
letterpaper, top=1in, bottom=1in, left=1in, right=1in]{geometry}
\hypersetup{urlcolor=blue, citecolor=blue, linkcolor=blue}
\usepackage{algorithm}
\usepackage{algorithmic}

\usepackage{bm}
\allowdisplaybreaks

\newtheorem{theorem}{Theorem}[section]

\theoremstyle{definition}

\newtheorem{remark}[theorem]{Remark}
\newtheorem{assumption}[theorem]{Assumption}

\newcommand{\R}{\mathbb{R}}
\newcommand{\E}{\mathbb{E}}
\newcommand{\N}{\mathcal{N}}
\newcommand{\M}{\mathcal{M}}
\newcommand{\D}{\mathcal{D}}
\newcommand{\dd}{\mathrm{d}}

\journal{Journal Name}

\begin{document}

\begin{frontmatter}

\title{Training-free score-based diffusion for parameter-dependent stochastic dynamical systems}

\author[ornl]{Minglei Yang\corref{cor1}}
\ead{yangm@ornl.gov}

\author[utk]{Sicheng He}
\ead{sicheng@utk.edu}

\cortext[cor1]{Corresponding author}

\affiliation[ornl]{organization={Fusion Energy Division, Oak Ridge National Laboratory},
            city={Oak Ridge},
            state={TN},
            country={USA}}

\affiliation[utk]{organization={Department of Mechanical and Aerospace Engineering, University of Tennessee},
            city={Knoxville},
            state={TN},
            country={USA}}


\begin{abstract}
Simulating parameter-dependent stochastic differential equations (SDEs) presents significant computational challenges, as separate high-fidelity simulations are typically required for each parameter value of interest. Despite the success of machine learning methods in learning SDE dynamics, existing approaches either require expensive neural network training for score function estimation or lack the ability to handle continuous parameter dependence. We present a training-free conditional diffusion model framework for learning stochastic flow maps of parameter-dependent SDEs, where both drift and diffusion coefficients depend on physical parameters. The key technical innovation is a joint kernel-weighted Monte Carlo estimator that approximates the conditional score function using trajectory data sampled at discrete parameter values, enabling interpolation across both state space and the continuous parameter domain. Once trained, the resulting generative model produces sample trajectories for any parameter value within the training range without retraining, significantly accelerating parameter studies, uncertainty quantification, and real-time filtering applications. The performance of the proposed approach is demonstrated via three numerical examples of increasing complexity, showing accurate approximation of conditional distributions across varying parameter values.
\end{abstract}

\begin{keyword}
Score-based diffusion models \sep parameter-dependent SDEs \sep stochastic flow maps \sep generative models \sep supervised learning
\end{keyword}

\tnotetext[fn1]{{\bf Notice}: 
This manuscript has been authored by UT-Battelle, LLC, under contract DE-AC05-00OR22725 with the US Department of Energy (DOE). The US government retains and the publisher, by accepting the article for publication, acknowledges that the US government retains a nonexclusive, paid-up, irrevocable, worldwide license to publish or reproduce the published form of this manuscript, or allow others to do so, for US government purposes. DOE will provide public access to these results of federally sponsored research in accordance with the DOE Public Access Plan.}

\end{frontmatter}

\section{Introduction}
Stochastic differential equations (SDEs) are fundamental mathematical models for describing dynamical systems subject to random fluctuations \cite{oksendal2003stochastic, khasminskii2012stochastic}. They arise naturally in diverse scientific and engineering applications, including financial modeling and various physical systems \cite{platen2010numerical, sobczyk2013stochastic}. In many of these applications, the drift and diffusion coefficients of the SDE depend on physical parameters that may vary across different operating conditions or experimental settings, such as diffusivity in transport processes or collision frequency in plasma physics. Understanding how these systems behave across a range of parameters is critical for uncertainty quantification (UQ), sensitivity analysis, and the design of reliable engineering systems~\cite{grigoriu2012stochastic}.

The mathematical significance of SDEs lies in their ability to provide both a particle-based description through stochastic trajectories and a continuum description via the associated Fokker--Planck partial differential equation~\cite{kolobov2003fokker, peeters2008fokker, yang2026efficient}. However, numerical solutions face substantial challenges in both formulations. Traditional particle-based methods, such as the Euler--Maruyama scheme and its high-order variants \cite{yuan2004convergence, piggott2016geometric}, can become computationally prohibitive when long-time simulations or large ensembles of trajectories are required. This computational burden is further exacerbated when parameter studies are needed, as separate high-fidelity simulations must be performed for each parameter value of interest. Meanwhile, PDE-based approaches~\cite{liu2004numerical, patie2008first} suffer from numerical instability, poor scalability with dimensions, and algorithmic challenges in parallelization.

Recently, machine learning (ML) approaches have emerged as promising alternatives for learning surrogate models of stochastic dynamics. Once trained, these models enable fast inference at new parameter values without requiring expensive repeated simulations, making them particularly attractive for parameter studies, uncertainty quantification, and integration with filtering frameworks such as differentiable Kalman filters~\cite{wu2025dkfnet}. These techniques span a broad spectrum, including Physics-informed neural networks~\cite{yang2020physics, chen2021solving}, Gaussian processes~\cite{archambeau2007gaussian, opper2019variational}, Flow Map Learning~\cite{chen2025due, chen2024learning}, polynomial approximations~\cite{werner2024sample}, and generative models such as Normalizing Flows~\cite{albergo2022building, yang2024pseudoreversible, urain2020imitationflow, yang2025conditional}.

Despite these successes, training generative models remains challenging due to mode collapse in GANs or architectural constraints in normalizing flows \cite{salimans2016improved, kobyzev2020normalizing}. Among various ML approaches, score-based diffusion models \cite{song2020score, nichol2021improved, yang2023diffusion} have demonstrated remarkable success by learning the score function to generate samples through a reverse-time process. A training-free conditional diffusion model was developed in \cite{liu2024training} for learning stochastic flow maps, where the exact score function admits a closed-form expression that can be estimated via Monte Carlo methods. Recent extensions have addressed SDEs in bounded domains via exit prediction networks~\cite{yang2025generative} and developed multi-fidelity approaches for amortized Bayesian inference~\cite{tatsuoka2025multi}. Error analysis in \cite{wang2026error} shows that the analytically tractable score function avoids accumulation of approximation errors, with bounds scaling as $\mathcal{O}(d)$ in $l^2$ and $\mathcal{O}(\log{d})$ in $l^{\infty}$.

In this work, we extend the training-free conditional diffusion framework to handle parameter-dependent SDEs where drift and diffusion coefficients depend on both the state and a physical parameter vector. The key technical innovation is a joint kernel-weighted Monte Carlo estimator that approximates the conditional score function using trajectory data sampled at discrete parameter values, enabling interpolation across both state space and the continuous parameter domain. We employ a two-stage procedure: the training-free score estimator first generates labeled training data via probability flow ODEs, then a feedforward neural network is trained via supervised regression on this data. This converts the challenging unsupervised generative modeling problem into a standard supervised learning task with training stability and architectural flexibility. The resulting framework learns a unified model that produces sample trajectories across different parameter values, enabling fast inference for parameter studies, uncertainty quantification, and real-time filtering.

The remainder of this paper is organized as follows. In Section~\ref{sec:probForm}, we introduce the parameter-dependent SDE formulation and the stochastic flow map representation. In Section~\ref{sec:Method}, we present the training-free conditional diffusion framework, including the generative model formulation, training data from Monte Carlo simulations, and training-free score estimation. The algorithm and implementation details are described in Section~\ref{sec:algorithm}. Numerical experiments are presented in Section~\ref{sec:Numeric}, followed by conclusions in Section~\ref{sec:conclude}.

\section{Problem formulation}\label{sec:probForm}

We consider dynamical systems governed by parameter-dependent SDEs.
The drift and diffusion coefficients depend on both the state and a parameter vector.
This section introduces the mathematical framework, including the SDE formulation and the stochastic flow map that characterizes one-step transitions.

\subsection{Parameter-dependent stochastic differential equations}

We consider the following $d$-dimensional parameter-dependent SDE:
\begin{equation}\label{eq:sde_general}
\dd X_t = a(X_t, \mu) \, \dd t + b(X_t, \mu) \, \dd W_t, \quad X_0 = x_0 \in \R^d,
\end{equation}
where $X_t \in \R^d$ is the state variable, $x_0 \in \R^d$ is the initial condition, $\mu \in \M \subset \R^{d_\mu}$ is a parameter vector belonging to a parameter domain $\M$, $a: \R^d \times \R^{d_\mu} \to \R^d$ is the drift coefficient, $b: \R^d \times \R^{d_\mu} \to \R^{d \times m}$ is the diffusion coefficient, and $W_t \in \R^m$ is an $m$-dimensional standard Brownian motion.
The parameter $\mu$ may represent various physical quantities depending on the application, e.g., diffusivity, or collision frequency \cite{wang2018computing, nguyen2016review}.

\begin{assumption}\label{assump:regularity}
We assume the following conditions hold for all $\mu \in \M$:
\begin{enumerate}
\item[(i)] The functions $a(\cdot, \mu)$ and $b(\cdot, \mu)$ satisfy the global Lipschitz condition: there exists $L > 0$ such that for all $x, y \in \R^d$,
\begin{equation}
\|a(x, \mu) - a(y, \mu)\| + \|b(x, \mu) - b(y, \mu)\|_F \leq L \|x - y\|.
\end{equation}
\item[(ii)] The functions $a(\cdot, \mu)$ and $b(\cdot, \mu)$ satisfy the linear growth condition: there exists $K > 0$ such that for all $x \in \R^d$,
\begin{equation}
\|a(x, \mu)\| + \|b(x, \mu)\|_F \leq K(1 + \|x\|).
\end{equation}
\item[(iii)] The functions $a$ and $b$ are continuous in $\mu$ for each fixed $x$.
\end{enumerate}
where $\|\cdot\|$ denotes the $l^2$ norm.
\end{assumption}

Under Assumption \ref{assump:regularity}, for each initial condition $x_0$ and parameter $\mu$, there exists a unique strong solution $\{X_t^{x_0, \mu}\}_{t \geq 0}$ to the SDE in Eq.~\eqref{eq:sde_general} \cite{oksendal2003stochastic}.
This ensures that the forward evolution is well-defined and that the conditional distributions we seek to learn are uniquely determined by the parameter value.

\subsection{The stochastic flow map}

To formulate the learning problem, we introduce a uniform temporal mesh
\begin{equation}\label{tmesh}
\mathcal{T} \coloneqq \{t_n : t_n = n\Delta t, \; n = 0, 1, \ldots, N_T \},
\end{equation}
where $T > 0$ is the final time, $N_T \in \mathbb{Z}^+$ is the number of time steps, and $\Delta t = T/N_T$ is the time step size.
On each time interval $[t_n, t_{n+1}]$, the SDE in Eq.~\eqref{eq:sde_general} can be written in the conditional form:
\begin{equation}\label{eq:flowmap}
X_{t_{n+1}}^{t_n,x} = x + \int_{t_n}^{t_{n+1}} a(X_s^{t_n,x}, \mu) \, \dd s + \int_{t_n}^{t_{n+1}} b(X_s^{t_n,x}, \mu) \, \dd W_s,
\end{equation}
where $X_{t_{n+1}}^{t_n,x}$ denotes the solution at time $t_{n+1}$ conditioned on $X_{t_n} = x$.
For notational simplicity, we denote $X_n \coloneqq X_{t_n}$ and define the \emph{stochastic flow map} $F_{\Delta t}$ such that
\begin{equation}\label{eq:flow_map}
X_{n+1} = X_n + F_{\Delta t}(X_n, \mu, \omega),
\end{equation}
where $\omega$ represents the randomness from the Brownian motion over the interval $[t_n, t_{n+1}]$.

\begin{remark}
The flow map $F_{\Delta t}(x, \mu, \omega)$ encapsulates the combined effects of drift, diffusion, and the parameter value $\mu$.
Unlike deterministic flow maps, this is a random variable for each fixed $(x, \mu)$ pair.
In terms of \eqref{eq:flowmap}, we have
\begin{equation}
F_{\Delta t}(x, \mu, \omega) = \int_{t_n}^{t_{n+1}} a(X_s^{t_n,x}, \mu) \, \dd s + \int_{t_n}^{t_{n+1}} b(X_s^{t_n,x}, \mu) \, \dd W_s.
\end{equation}
This representation makes explicit the dependence on both the drift and diffusion terms integrated over the time interval.
\end{remark}

The conditional distribution of $X_{n+1}$ given $X_n = x_n$ and parameter $\mu$, which we denote by
\begin{equation}\label{eq:cond_dist}
X_{n+1} \mid (X_n = x_n, \mu) \sim p(\cdot \mid x_n, \mu),
\end{equation}
is the transition kernel of the Markov process parameterized by $\mu$.
Learning this parameter-dependent conditional distribution $p(\cdot \mid x_n, \mu)$ is the central objective of this work.
The key challenge is to construct a generative model that accurately captures this distribution across the entire parameter domain $\M$, enabling efficient sampling for any parameter value $\mu$ without repeated expensive simulations.

\section{Training-free conditional diffusion framework}\label{sec:Method}

Having defined the parameter-dependent SDE and its associated flow map, we now present the proposed method for learning the stochastic flow map.
The methodology consists of four main components: (1) the generative model formulation; (2) training data from Monte Carlo simulations; (3) a training-free score estimation approach; and (4) a supervised learning framework for training the generative model.
The approach extends the framework developed in \cite{liu2024training, liu2024diffusion} to the parameter-dependent setting.

The term ``training-free'' in this work refers specifically to the score function estimation: unlike standard diffusion models that train neural networks to approximate the score function via score-matching losses, our approach derives a closed-form expression that can be directly estimated via Monte Carlo methods using trajectory data. The final stochastic flow map $G_\theta$ is subsequently learned through supervised regression on the diffusion-generated labeled data. This two-stage procedure converts the challenging unsupervised generative modeling problem into a standard supervised learning task.

\subsection{Generative model for the stochastic flow map}\label{sec:generative_model}

To approximate the target distribution $p(\cdot \mid x_n, \mu)$, we aim to learn a parameterized generative model $G_\theta: \R^d \times \R^{d_\mu} \times \R^d \to \R^d$ such that
\begin{equation}\label{eq:generative_model}
\hat{X}_{n+1} = X_n + G_\theta(X_n, \mu, Z), \quad Z \sim \N(0, I_d),
\end{equation}
where $Z \in \R^d$ is a standard $d$-dimensional Gaussian random variable and $\theta \in \R^p$ denotes the neural network parameters with dimension $p$ determined by the network architecture.
The neural network $G_\theta$ serves as an approximation to the stochastic flow map $F_{\Delta t}$ defined in Eq.~\eqref{eq:flow_map}, replacing the abstract randomness $\omega$ with an explicit Gaussian latent variable $z$.
The distribution of $\hat{X}_{n+1}$ should match the true conditional distribution $p(\cdot \mid X_n, \mu)$ for all $(X_n, \mu) \in \R^d \times \M$.

The key challenge is that labeled training data of the form $(x_n, \mu, z, x_{n+1})$ are not directly available from Monte Carlo simulations: while we can observe state transitions $(x_n, x_{n+1})$ for each parameter $\mu$, the latent Gaussian variable $z$ that maps to each specific $x_{n+1}$ is unknown.
We address this challenge by using a score-based diffusion model to generate such labeled data, forming an augmented dataset $\D_{\text{aug}} = \{(x_n, \mu, z, \hat{x}_{n+1})\}$.

\subsection{Training data from Monte Carlo simulations}

We assume access to trajectory data generated by Monte Carlo simulations of the SDE \eqref{eq:sde_general} at various parameter values.
Specifically, let $N_\mu \in \mathbb{N}$ denote the number of parameter values and $N_s \in \mathbb{N}$ denote the number of trajectories per parameter.
For a collection of parameter values $\{\mu^{(k)}\}_{{k=1}}^{{N_\mu}} \subset \M$, we simulate $N_s$ independent trajectories for each parameter value, yielding the observed dataset
\begin{equation}\label{eq:dataset}
\D_{\rm obs} = \left\{ \left( x_n^{(i,k)}, \mu^{(k)}, x_{n+1}^{(i,k)} \right) : i = 1, \ldots, N_s, \; k = 1, \ldots, N_\mu, \; n = 0, \ldots, N_T - 1 \right\}.
\end{equation}
For notational convenience, we reorganize this dataset as a collection of state transition pairs:
\begin{equation}\label{eq:dataset_pairs}
\D_{\rm obs} = \left\{ (x_n^{(j)}, \mu^{(j)}, x_{n+1}^{(j)}) \right\}_{j=1}^{J},
\end{equation}
where $J = N_s \times N_\mu \times N_T$ is the total number of training pairs, $x_n^{(j)}$ and $x_{n+1}^{(j)}$ represent consecutive states, and $\mu^{(j)}$ is the associated parameter value.
The central task is to generate the augmented dataset $\D_{\text{aug}}$ from $\D_{\rm obs}$; the following subsections present a training-free diffusion approach to accomplish this.

\subsection{Overview of the diffusion-based approach}

We employ a score-based diffusion model to learn the parameter-dependent conditional distribution $p(x_{n+1} \mid x_n, \mu)$, which describes the transition probability from state $x_n$ to $x_{n+1}$ for each fixed parameter value $\mu$.
The key idea is to construct a diffusion process that transforms this conditional distribution into a standard Gaussian, and then reverse this process to generate samples.
The framework builds upon score-based diffusion theory \cite{anderson1982reverse, song2020score}, which we extend to handle parameter-dependent conditional distributions.

The approach consists of two stages:
\begin{enumerate}
\item[(i)] \emph{Forward diffusion}: Transform the conditional distribution $p(\cdot \mid x_n, \mu)$ into a standard Gaussian $\N(0, I_d)$ via a diffusion process on an artificial time domain $\tau \in [0, 1]$.
\item[(ii)] \emph{Reverse-time ODE}: Run the reverse process deterministically to map latent Gaussian samples $z \sim \N(0, I_d)$ back to samples from $p(\cdot \mid x_n, \mu)$.
\end{enumerate}

The reverse-time process is governed by the \emph{probability flow ODE}:
\begin{equation}\label{eq:reverse_ode}
\dd Z_\tau^{x_n,\mu} = \left[ f(\tau) Z_\tau^{x_n,\mu} - \frac{1}{2} g^2(\tau) S(Z_\tau^{x_n,\mu}, \tau; x_n, \mu) \right] \dd \tau,
\end{equation}
where $f(\tau) = -1/(1-\tau)$ and $g(\tau) = \sqrt{(1+\tau)/(1-\tau)}$ are the drift and diffusion coefficients of the variance-preserving (VP) forward process (see~\ref{app:diffusion} for derivation), and $S(z, \tau; x_n, \mu)$ is the \emph{parameter-dependent conditional score function}:
\begin{equation}\label{eq:score_def}
S(z, \tau; x_n, \mu) \coloneqq \nabla_z \log Q(Z_\tau^{x_n,\mu} = z \mid x_n, \mu),
\end{equation}
where $Q(Z_\tau^{x_n,\mu} = z \mid x_n, \mu)$ denotes the marginal density of the diffusion state at time $\tau$ conditioned on $(x_n, \mu)$.

The diffusion process is applied to the \emph{displacement} $\Delta X_{n+1} \coloneqq X_{n+1} - X_n$ rather than the absolute state $X_{n+1}$.
The ODE \eqref{eq:reverse_ode} thus establishes a deterministic mapping from $Z_1 \sim \N(0, I_d)$ to $Z_0 \sim p_\Delta(\cdot \mid x_n, \mu)$, where $p_\Delta(\cdot \mid x_n, \mu)$ denotes the conditional distribution of the displacement.
This deterministic nature is crucial for generating labeled training data, as it establishes a one-to-one correspondence between each latent Gaussian variable $z$ and its corresponding displacement $\Delta x_{n+1}$.
Complete derivations of the forward and reverse diffusion processes are provided in~\ref{app:diffusion}.

\subsection{Training-free score estimation}

A key contribution of \cite{liu2024training, liu2024diffusion} is the derivation of a closed-form expression for the conditional score function that can be estimated directly from data without training a neural network.
We extend this approach to the parameter-dependent setting.

The score function in Eq.~\eqref{eq:score_def} can be expressed as a weighted average over the displacement distribution $p_\Delta(\Delta x \mid x_n, \mu)$ (see~\ref{app:score} for the complete derivation):
\begin{equation}\label{eq:score_weighted}
S(z, \tau; x_n, \mu) = \int_{\R^d} \left( -\frac{z - \alpha_\tau \Delta x}{\beta_\tau^2} \right) w_\tau(z, \Delta x; x_n, \mu) \, p_\Delta(\Delta x \mid x_n, \mu) \, \dd (\Delta x),
\end{equation}
where $\alpha_\tau = 1 - \tau$ and $\beta_\tau^2 = \tau$ are the VP schedule scaling functions, and $\Delta x = x_{n+1} - x_n$ denotes the displacement.
The weight function $w_\tau(z, \Delta x; x_n, \mu)$ favors displacement values more likely to have produced the diffusion state $z$; see~\ref{app:score} for its explicit form.
This representation reveals that the score function is interpretable as a weighted average of ``local scores'' $-(z - \alpha_\tau \Delta x)/\beta_\tau^2$ over all possible displacement values.

Since the displacement distribution $p_\Delta(\Delta x \mid x_n, \mu)$ is unknown, we approximate the score function $S$ in Eq.~\eqref{eq:score_weighted} via Monte Carlo estimation.
In principle, if we had displacement samples $\{\Delta x^{(\ell)}\}_{\ell=1}^N$ drawn exactly from $p_\Delta(\cdot \mid x_n, \mu)$, we could approximate the score by replacing the integral with a sum and using the theoretical weights $w_\tau$ from Eq.~\eqref{eq:score_weighted}.
However, our dataset $\D_{\rm obs}$ contains samples from various $(x_n, \mu)$ pairs, not necessarily at the exact query point.

To address this, we select the $N$ nearest neighbors of $(x_n, \mu)$ from $\D_{\rm obs}$ in the joint space, denoted $\{(x_n^{(j_\ell)}, \mu^{(j_\ell)}, x_{n+1}^{(j_\ell)})\}_{\ell=1}^N$, and compute the corresponding displacements $\Delta x^{(j_\ell)} = x_{n+1}^{(j_\ell)} - x_n^{(j_\ell)}$.
The score function is then approximated by:
\begin{equation}\label{eq:score_mc}
\bar{S}(z, \tau; x_n, \mu) \coloneqq \sum_{\ell=1}^{N} \left( -\frac{z - \alpha_\tau \Delta x^{(j_\ell)}}{\beta_\tau^2} \right) \bar{w}_\tau^\ell,
\end{equation}
where $\bar{w}_\tau^\ell$ are the normalized Monte Carlo weights that approximate the theoretical weights $w_\tau$ in Eq.~\eqref{eq:score_weighted}.
Since the neighbor samples are not exactly at $(x_n, \mu)$, we extend the theoretical weight (which only involves the Gaussian transition density $Q(z \mid \Delta x)$) by incorporating kernel density estimation for the proximity in both $x_n$ and $\mu$:
\begin{equation}\label{eq:weight_mc}
\bar{w}_\tau^\ell \propto \underbrace{\exp\left( -\frac{\|z - \alpha_\tau \Delta x^{(j_\ell)}\|^2}{2\beta_\tau^2}\right)}_{\text{diffusion weight } Q(z \mid \Delta x^{(j_\ell)})} \cdot \underbrace{\exp\left(- \frac{\|x_n - x_n^{(j_\ell)}\|^2}{2\nu_x^2}\right)}_{\text{spatial kernel}} \cdot \underbrace{\exp\left(- \frac{\|\mu - \mu^{(j_\ell)}\|^2}{2\nu_\mu^2} \right)}_{\text{parameter kernel}},
\end{equation}
with normalization $\sum_{\ell=1}^N \bar{w}_\tau^\ell = 1$.
The first term corresponds to the theoretical weight $w_\tau$ derived in Eq.~\ref{app:score}, while the spatial and parameter kernels account for the fact that neighbor samples originate from nearby but not identical $(x_n, \mu)$ values.
Here, $\nu_x > 0$ and $\nu_\mu > 0$ are bandwidth parameters controlling the influence of spatial and parameter proximity, respectively.
This joint kernel weighting approach naturally handles interpolation in both the state variable $x_n$ and the parameter $\mu$ simultaneously, enabling generalization beyond the discrete training samples.
When $\tau$ is small, $\beta_\tau^2 \approx 0$ causes the diffusion weights to become sharply peaked; in practice, we add a small regularization $\delta > 0$ to $\beta_\tau^2$ for numerical stability.

\section{Algorithm and implementation}\label{sec:algorithm}

\begin{algorithm}[htbp]
\caption{Training-free conditional diffusion for parameter-dependent flow maps}
\label{alg:main}
\begin{algorithmic}[1]
\REQUIRE Observed dataset $\D_{\rm obs} = \{(x_n^{(j)}, \mu^{(j)}, x_{n+1}^{(j)})\}_{j=1}^{J}$, number of labeled samples $M$, ODE time steps $N_\tau$, number of neighbors $N$
\ENSURE Trained generative model $G_\theta$

\STATE $\D_{\text{aug}} \leftarrow \emptyset$ \COMMENT{Initialize augmented dataset}
\FOR{$m = 1, \ldots, M$}
\STATE $(x_n^{(m)}, \mu^{(m)}) \sim \D_{\rm obs}$ \COMMENT{Sample query point}
\STATE $z^{(m)} \sim \N(0, I_d)$ \COMMENT{Sample latent variable}
\STATE $Z_{N_\tau} \leftarrow z^{(m)}$ \COMMENT{Initialize ODE}
\FOR{$k = N_\tau, N_\tau - 1, \ldots, 1$}
\STATE $\tau_k \leftarrow k / N_\tau$, \quad $\Delta\tau \leftarrow 1/N_\tau$ \COMMENT{Diffusion time}
\STATE $\{(x_n^{(j_\ell)}, \mu^{(j_\ell)}, x_{n+1}^{(j_\ell)})\}_{\ell=1}^N \leftarrow \textsc{NearestNeighbor}(x_n^{(m)}, \mu^{(m)}, N, \D_{\rm obs})$ \COMMENT{Neighbors in $(x,\mu)$}
\STATE $\Delta x^{(j_\ell)} \leftarrow x_{n+1}^{(j_\ell)} - x_n^{(j_\ell)}$ for $\ell = 1, \ldots, N$ \COMMENT{Compute displacements}
\STATE $\bar{S} \leftarrow \sum_{\ell=1}^{N} \bar{w}_{\tau_k}^\ell \cdot \left( -\frac{Z_k - (1-\tau_k) \Delta x^{(j_\ell)}}{\tau_k} \right)$ \COMMENT{Score \eqref{eq:score_mc}, weights \eqref{eq:weight_mc} with $(x_n, \mu)$ kernels}
\STATE $Z_{k-1} \leftarrow Z_k - \Delta\tau \left[ f(\tau_k) Z_k - \frac{1}{2} g^2(\tau_k) \bar{S} \right]$ \COMMENT{Euler step \eqref{eq:reverse_ode}}
\ENDFOR
\STATE $\hat{x}_{n+1}^{(m)} \leftarrow x_n^{(m)} + Z_0$ \COMMENT{Displacement $Z_0$ plus initial state}
\STATE $\D_{\text{aug}} \leftarrow \D_{\text{aug}} \cup \{(x_n^{(m)}, \mu^{(m)}, z^{(m)}, \hat{x}_{n+1}^{(m)})\}$ \COMMENT{Labeled data \eqref{eq:labeled_data}}
\ENDFOR

\STATE $G_\theta \leftarrow \text{InitNN}()$ \COMMENT{Initialize neural network}
\REPEAT
\STATE $\{(x_n^{(i)}, \mu^{(i)}, z^{(i)}, \hat{x}_{n+1}^{(i)})\}_i \subset \D_{\text{aug}}$ \COMMENT{Sample mini-batch}
\STATE $\theta \leftarrow \theta - \eta \nabla_\theta \sum_i \| G_\theta(x_n^{(i)}, \mu^{(i)}, z^{(i)}) - (\hat{x}_{n+1}^{(i)} - x_n^{(i)}) \|^2$ \COMMENT{Gradient step \eqref{eq:mse_loss}}
\UNTIL{$\mathcal{L}_{\text{val}}(\theta)$ converges} \COMMENT{Early stopping}
\RETURN $G_\theta$
\end{algorithmic}
\end{algorithm}

The algorithm first generates labeled data by solving the probability flow ODE in Eq.~\eqref{eq:reverse_ode}, which establishes a deterministic mapping from latent space $Z_1 \sim \N(0, I_d)$ to the target distribution:
\begin{equation}\label{eq:labeled_data}
\D_{\text{aug}} = \left\{ (x_n^{(m)}, \mu^{(m)}, z^{(m)}, \hat{x}_{n+1}^{(m)}) \right\}_{m=1}^{M}.
\end{equation}
The neural network $G_\theta$ is then trained via supervised learning to minimize the mean squared error (MSE) loss $\mathcal{L}(\theta)$ defined as
\begin{equation}\label{eq:mse_loss}
\mathcal{L}(\theta) = \frac{1}{M} \sum_{m=1}^{M} \left\| G_\theta(x_n^{(m)}, \mu^{(m)}, z^{(m)}) - (\hat{x}_{n+1}^{(m)} - x_n^{(m)}) \right\|^2,
\end{equation}
where training uses early stopping based on the validation loss $\mathcal{L}_{\text{val}}(\theta)$, computed analogously on a held-out subset of $\D_{\text{aug}}$.
The main computational cost is the nearest neighbor search during labeled data generation; our implementation uses $k$-d trees in \cite{bentley1975kdtree} for efficient neighbor queries.
The ODE solving requires $N_\tau$ evaluations of the score function per labeled sample.

\section{Numerical experiments}\label{sec:Numeric}

In this section, we present numerical experiments to validate the proposed method across three test problems of increasing complexity.
Example~1 considers Brownian motion with parameter-dependent drift ($d = 1$, $d_\mu = 1$), where $\mu$ controls the drift coefficient, providing a validation case with known analytical solutions.
Example~2 studies an Ornstein-Uhlenbeck (OU) process with parameter-dependent drift and diffusion ($d = 1$, $d_\mu = 1$), where $\mu$ appears in both coefficients and the stationary variance exhibits non-monotonic dependence on $\mu$.
Example~3 considers a two-dimensional OU process with rotation ($d = 2$, $d_\mu = 1$), demonstrating the method's capability for multi-dimensional state spaces with coupled dynamics.

All experiments are implemented in PyTorch \cite{paszke2019pytorch} and executed on a workstation with an NVIDIA GPU.
The neural network $G_\theta$ uses a fully-connected architecture with tanh activation functions, trained using the Adam optimizer \cite{kingma2014adam} with learning rate $0.001$ and batch size $1024$.
Training employs early stopping with patience of $50$ epochs based on validation loss, using a $90\%/10\%$ train/validation split.
Specific hyperparameters (network size, number of labeled samples $M$, ODE steps $N_\tau$, neighbors $N$, and bandwidth parameters $\nu_x$, $\nu_\mu$) are tuned per example and reported in each subsection.
Our implementation uses $k$-d trees \cite{bentley1975kdtree} for efficient neighbor queries during label generation.

\begin{remark}[Reproducibility]
The source code for all numerical experiments is publicly available at \url{https://github.com/mlmathphy/Diffusion_parameter_SDE}.
All results presented in this section are fully reproducible using the provided repository.
\end{remark}

\subsection{Example 1: Brownian motion with parameter-dependent drift}

Consider the general SDE \eqref{eq:sde_general} with $d = m = 1$ and coefficients:
\begin{equation}\label{eq:ex1_coeffs}
a(x, \mu) = \mu, \quad b(x, \mu) = 1,
\end{equation}
where $\mu \in [-1, 1]$ is the drift parameter.
This yields the one-dimensional SDE:
\begin{equation}\label{eq:ex1_sde}
\dd X_t = \mu \, \dd t + \dd W_t, \quad X_0 = x_0,
\end{equation}
where $W_t$ is a standard Brownian motion.
This example serves as a validation case since exact analytical solutions are available.

\subsubsection{Exact solutions}

For the SDE in Eq.~\eqref{eq:ex1_sde}, the exact solution is
\begin{equation}
X_t = X_0 + \mu t + W_t.
\end{equation}
The one-step conditional distribution $p(X_{n+1} \mid X_n = x, \mu)$ is Gaussian:
\begin{equation}\label{eq:ex1_exact_cond}
p(X_{n+1} \mid X_n = x, \mu) = \N\left(x + \mu \Delta t, \, \Delta t\right),
\end{equation}
with conditional mean and variance:
\begin{equation}
\E[X_{n+1} \mid X_n = x, \mu] = x + \mu \Delta t, \quad \text{Var}[X_{n+1} \mid X_n = x, \mu] = \Delta t.
\end{equation}
The conditional mean depends linearly on both the initial state $x$ and the drift parameter $\mu$, while the variance depends only on the time step $\Delta t$.

For the terminal distribution at time $T$ with a Gaussian initial condition $X_0 \sim \N(m_0, \sigma_0^2)$, the exact distribution is also Gaussian:
\begin{equation}\label{eq:ex1_terminal}
X_T \sim \N\left(m_0 + \mu T, \, \sigma_0^2 + T\right).
\end{equation}
This follows from the fact that $X_T = X_0 + \mu T + W_T$, where $X_0$ and $W_T$ are independent Gaussian random variables.

\subsubsection{Training procedure}

We generate training data by simulating the SDE in Eq.~\eqref{eq:ex1_sde} using the Euler-Maruyama scheme with a fine time step $\delta t = 0.001$.
The training dataset consists of consecutive state pairs $(X_n, X_{n+1})$ sampled at $N_\mu = 21$ uniformly spaced parameter values $\mu \in [-1, 1]$.
For each parameter value, we simulate $N_s = 5000$ independent trajectories over the time interval $[0, T]$ with $T = 1.0$ and recording time step $\Delta t = 0.1$.
The initial conditions are sampled uniformly from $[-5, 5]$, yielding a total of $J = 1{,}050{,}000$ training pairs.

Following Algorithm \ref{alg:main}, we generate $M = 50{,}000$ labeled training samples using the training-free conditional diffusion model with $N = 2000$ nearest neighbors based on the combined distance in $(x, \mu)$ space.
The probability flow ODE in Eq.~\eqref{eq:reverse_ode} is solved using the explicit Euler scheme with $N_\tau = 1{,}000$ time steps.
The bandwidth parameters for the Gaussian kernel weights are set to $\nu_x = 1.0$ for spatial proximity and $\nu_\mu = 0.5$ for parameter proximity.
The generative model $G_\theta(x, \mu, z)$ is parameterized as a fully-connected neural network with input dimension $3$, three hidden layers with $128$ neurons each, and output dimension $1$.
The network predicts the scaled displacement $(\hat{x}_{n+1} - x_n) \cdot c_{\text{scale}}$ with $c_{\text{scale}} = 3.0$; this scaling normalizes the network output to approximately unit variance, improving training stability and convergence.

\subsubsection{Results}

\begin{figure}[htbp]
\centering
\includegraphics[width=0.9\textwidth]{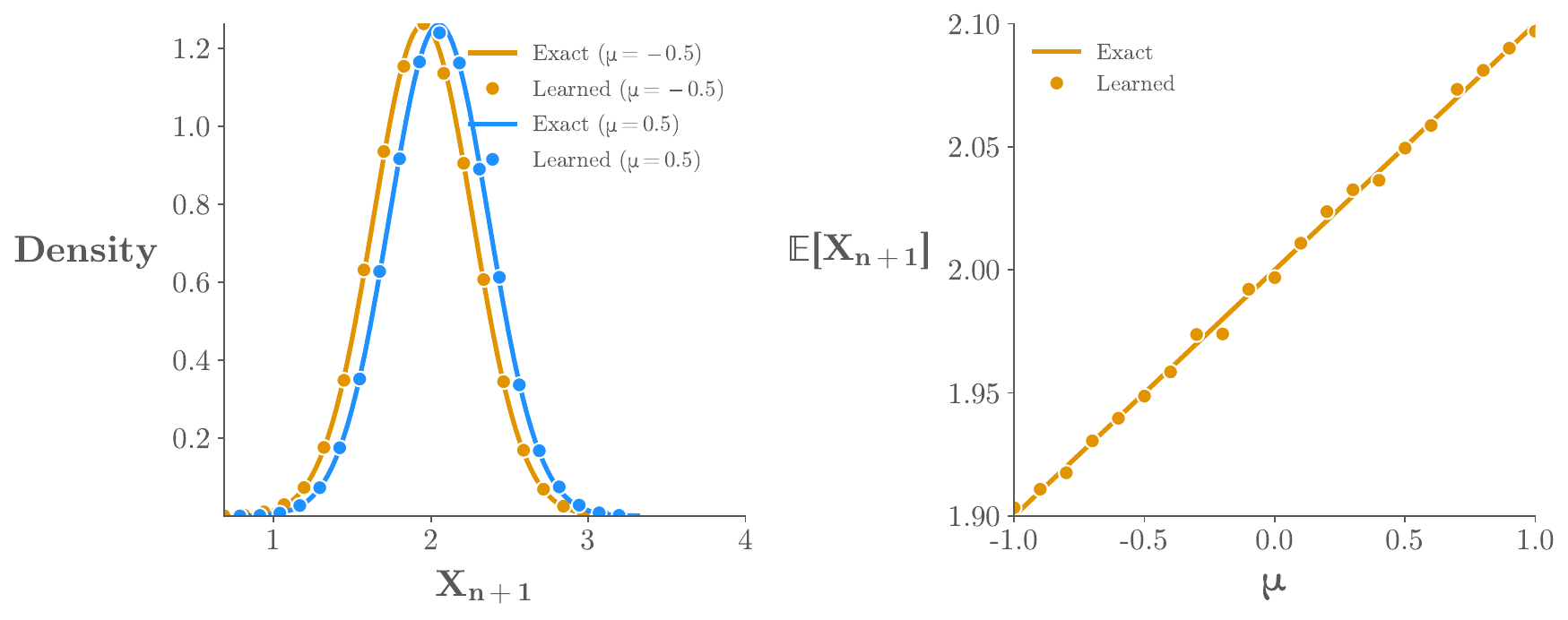}
\caption{Example 1: Conditional distribution $p(X_{n+1} \mid X_n = 2, \mu)$. Left: PDF comparison for $\mu = -0.5$ and $\mu = 0.5$; solid lines show the exact Gaussian density \eqref{eq:ex1_exact_cond}, markers show learned estimates. Right: Conditional mean $\E[X_{n+1} \mid X_n = 2, \mu]$ vs.\ $\mu$; solid line is the exact formula $x + \mu \Delta t$, markers are learned estimates.}
\label{fig:ex1_conditional}
\end{figure}

Figure \ref{fig:ex1_conditional} shows the learned conditional distribution compared with the exact solution in Eq.~\eqref{eq:ex1_exact_cond}.
The left panel displays the probability density function $p(X_{n+1} \mid X_n = 2, \mu)$ for two parameter values $\mu = -0.5$ and $\mu = 0.5$.
The exact conditional means are $2 + (-0.5)(0.1) = 1.95$ and $2 + (0.5)(0.1) = 2.05$, respectively, with conditional standard deviation $\sqrt{\Delta t} = \sqrt{0.1} \approx 0.316$ for both cases.
The learned distributions (markers), estimated using kernel density estimation from $50{,}000$ generated samples, closely match the exact Gaussian distributions (solid lines).
The right panel shows the conditional mean $\E[X_{n+1} \mid X_n = 2, \mu]$ as a function of $\mu$.
The learned means (markers), computed from $10{,}000$ samples per parameter value, accurately capture the linear dependence $x + \mu \Delta t$ predicted by the exact formula (solid line).
Quantitatively, the maximum absolute error in the conditional mean is $0.006$ (mean absolute error $0.002$), and the maximum relative error in the conditional variance is $2.5\%$ (mean relative error $1.1\%$).

\begin{figure}[htbp]
\centering
\includegraphics[width=0.9\textwidth]{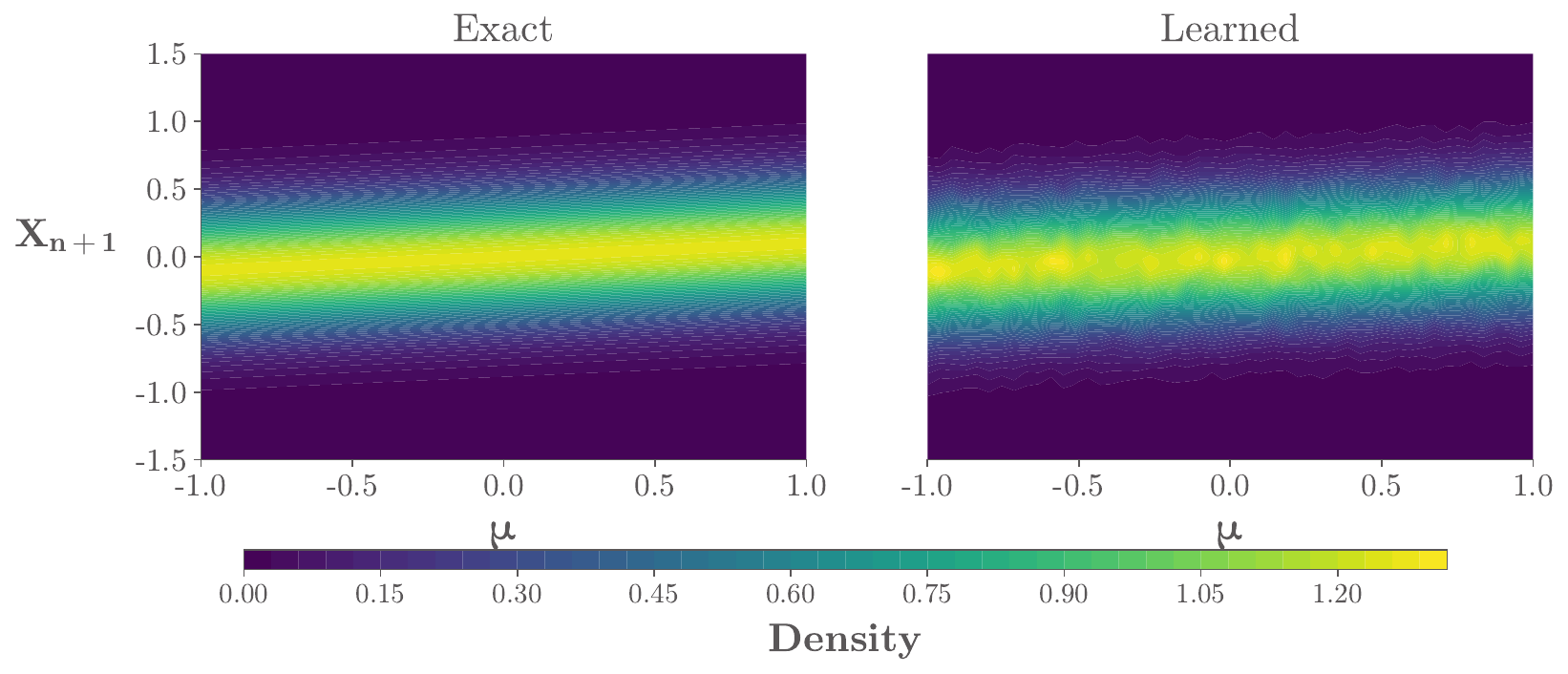}
\caption{Example 1: Heatmap of conditional distribution $p(X_{n+1} \mid X_n = 0, \mu)$ over the parameter range $\mu \in [-1, 1]$. Left: Exact distribution from \eqref{eq:ex1_exact_cond}. Right: Learned distribution. The heatmaps are constructed using $5{,}000$ samples per $\mu$ value. The learned model captures the linear shift in mean with $\mu$.}
\label{fig:ex1_heatmap}
\end{figure}

Figure \ref{fig:ex1_heatmap} presents a heatmap comparison of the conditional distribution $p(X_{n+1} \mid X_n = 0, \mu)$ over the entire parameter range.
The left panel shows the exact distribution computed from Eq.~\eqref{eq:ex1_exact_cond}, and the right panel shows the learned distribution.
The learned model accurately captures the shift in the distribution mean as $\mu$ varies from $-1$ to $1$.
Notably, the model successfully interpolates to parameter values not included in the training set (which used $N_\mu = 21$ discrete values), demonstrating the method's ability to generalize across the continuous parameter space.

\begin{figure}[htbp]
\centering
\includegraphics[width=0.9\textwidth]{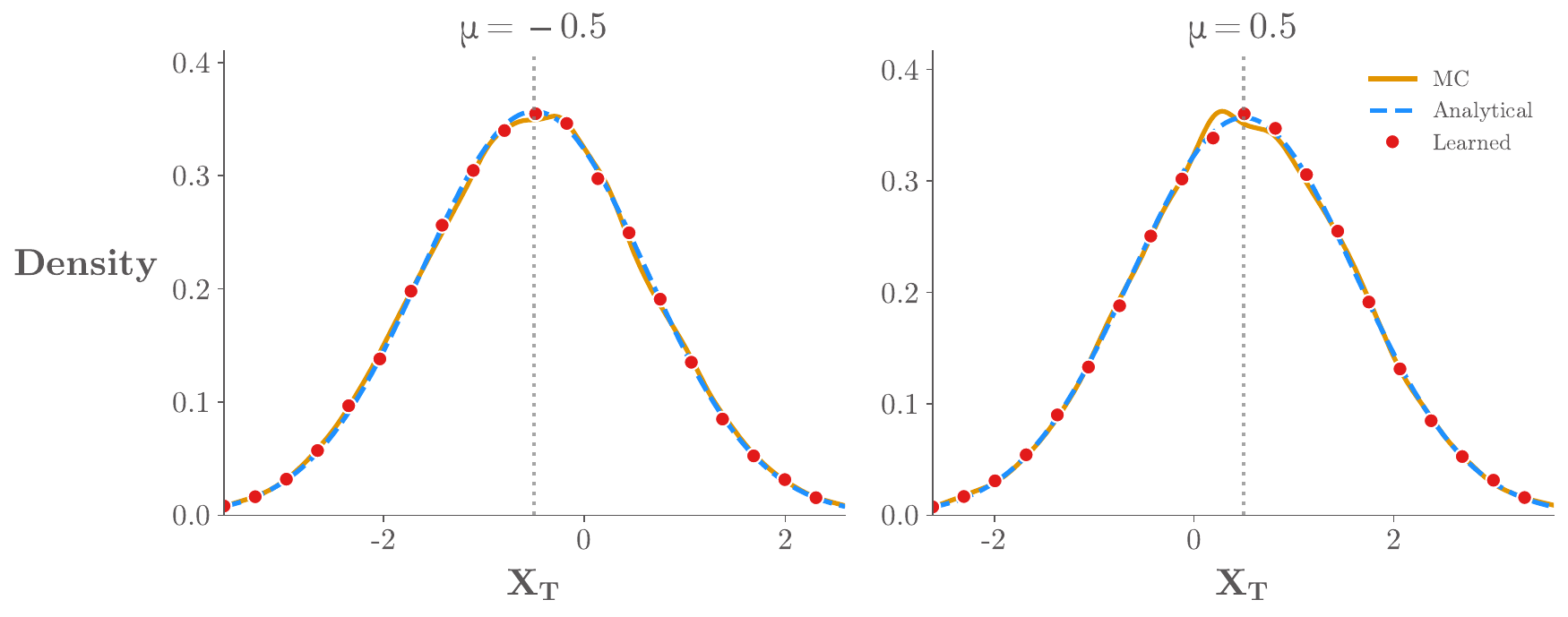}
\caption{Example 1: Terminal distribution $p(X_T)$ at $T = 1.0$ with Gaussian initial distribution $X_0 \sim \N(0, 0.25)$, obtained via $10$ iterative applications of the learned one-step flow map. Left panel: $\mu = -0.5$. Right panel: $\mu = 0.5$. Solid lines: Monte Carlo ground truth ($50{,}000$ samples); dashed lines: analytical formula \eqref{eq:ex1_terminal}; markers: learned method ($50{,}000$ generated trajectories). Vertical dotted lines indicate the analytical mean $m_0 + \mu T$.}
\label{fig:ex1_terminal}
\end{figure}

To demonstrate the capability of multi-step prediction, Figure \ref{fig:ex1_terminal} shows the terminal distribution $p(X_T)$ at $T = 1.0$ starting from a non-delta initial distribution $X_0 \sim \N(0, 0.25)$.
Multi-step trajectories are generated by iteratively applying the learned one-step flow map with $N_{\text{steps}} = T/\Delta t = 10$ steps.
The figure compares three approaches: (i) Monte Carlo ground truth using $50{,}000$ samples from exact SDE simulation (solid lines), (ii) the analytical formula \eqref{eq:ex1_terminal} (dashed lines), and (iii) our learned method using $50{,}000$ generated trajectories (markers).
The vertical dotted lines indicate the analytical mean $m_0 + \mu T$, which equals $-0.5$ for $\mu = -0.5$ (left panel) and $0.5$ for $\mu = 0.5$ (right panel).
According to \eqref{eq:ex1_terminal}, the terminal variance is $\sigma_0^2 + T = 0.25 + 1.0 = 1.25$ for both cases.
The learned distributions closely match both the Monte Carlo reference and the analytical solution, validating the accuracy of the learned flow map for long-time prediction.
Quantitatively, the terminal mean errors are below $0.006$ and the terminal standard deviation errors are below $0.007$ for both parameter values.

\subsection{Example 2: Ornstein-Uhlenbeck process with parameter-dependent drift and diffusion}

To demonstrate the method's capability when the parameter appears in both drift and diffusion coefficients, we consider the general SDE \eqref{eq:sde_general} with $d = m = 1$ and coefficients:
\begin{equation}\label{eq:ex2_coeffs}
a(x, \mu) = -\mu x, \quad b(x, \mu) = \sqrt{1 + \mu^2},
\end{equation}
where $\mu \in [0.5, 2]$ controls both the mean-reversion rate and the diffusion intensity.
This yields the Ornstein-Uhlenbeck process:
\begin{equation}\label{eq:ex2_sde}
\dd X_t = -\mu X_t \, \dd t + \sqrt{1 + \mu^2} \, \dd W_t, \quad X_0 = x_0.
\end{equation}
This example is particularly interesting because the stationary variance exhibits non-monotonic dependence on $\mu$.

\subsubsection{Exact solutions}

For the SDE \eqref{eq:ex2_sde}, the exact solution is
\begin{equation}
X_t = X_0 e^{-\mu t} + \sqrt{1 + \mu^2} \int_0^t e^{-\mu(t-s)} \, \dd W_s.
\end{equation}
The one-step conditional distribution $p(X_{n+1} \mid X_n = x, \mu)$ is Gaussian:
\begin{equation}\label{eq:ex2_exact_cond}
p(X_{n+1} \mid X_n = x, \mu) = \N\left(x e^{-\mu \Delta t}, \, \frac{1 + \mu^2}{2\mu}\left(1 - e^{-2\mu \Delta t}\right)\right),
\end{equation}
with conditional mean and variance:
\begin{align}
\E[X_{n+1} \mid X_n = x, \mu] &= x e^{-\mu \Delta t}, \\
\text{Var}[X_{n+1} \mid X_n = x, \mu] &= \frac{1 + \mu^2}{2\mu}\left(1 - e^{-2\mu \Delta t}\right).
\end{align}
The conditional mean exhibits exponential decay toward zero, while the variance depends on both $\mu$ and $\Delta t$ through a more complex expression.

The stationary distribution is $\N(0, \sigma_\infty^2(\mu))$ with variance:
\begin{equation}\label{eq:ex2_stationary}
\sigma_\infty^2(\mu) = \frac{1 + \mu^2}{2\mu}.
\end{equation}
Notably, this variance is non-monotonic in $\mu$: it achieves its minimum value of $1$ at $\mu = 1$, and increases for both $\mu < 1$ and $\mu > 1$.

For the terminal distribution at time $T$ with a Gaussian initial condition $X_0 \sim \N(m_0, \sigma_0^2)$, the exact distribution is:
\begin{equation}\label{eq:ex2_terminal}
X_T \sim \N\left(m_0 e^{-\mu T}, \, \sigma_0^2 e^{-2\mu T} + \frac{1 + \mu^2}{2\mu}\left(1 - e^{-2\mu T}\right)\right).
\end{equation}
This expression combines the exponential decay of the initial condition with the approach to the stationary distribution, and will be used to validate multi-step predictions.

\subsubsection{Training procedure}

We generate training data using the Euler-Maruyama scheme with fine time step $\delta t = 0.001$.
The dataset consists of state pairs $(X_n, X_{n+1})$ at $N_\mu = 21$ uniformly spaced values $\mu \in [0.5, 2]$.
For each parameter value, we simulate $N_s = 5000$ trajectories with recording time step $\Delta t = 0.1$ and total time $T = 1.0$.
Initial conditions are sampled from the stationary distribution \eqref{eq:ex2_stationary} for each $\mu$, yielding a total of $J = 1{,}050{,}000$ training pairs.
Following Algorithm \ref{alg:main}, we generate $M = 20{,}000$ labeled samples using the training-free conditional diffusion model with $N = 1000$ nearest neighbors and $N_\tau = 500$ ODE time steps.
The bandwidth parameters are $\nu_x = 1.0$ and $\nu_\mu = 0.3$, and the neural network architecture remains the same as in Example 1.

\subsubsection{Results}

\begin{figure}[htbp]
\centering
\includegraphics[width=0.9\textwidth]{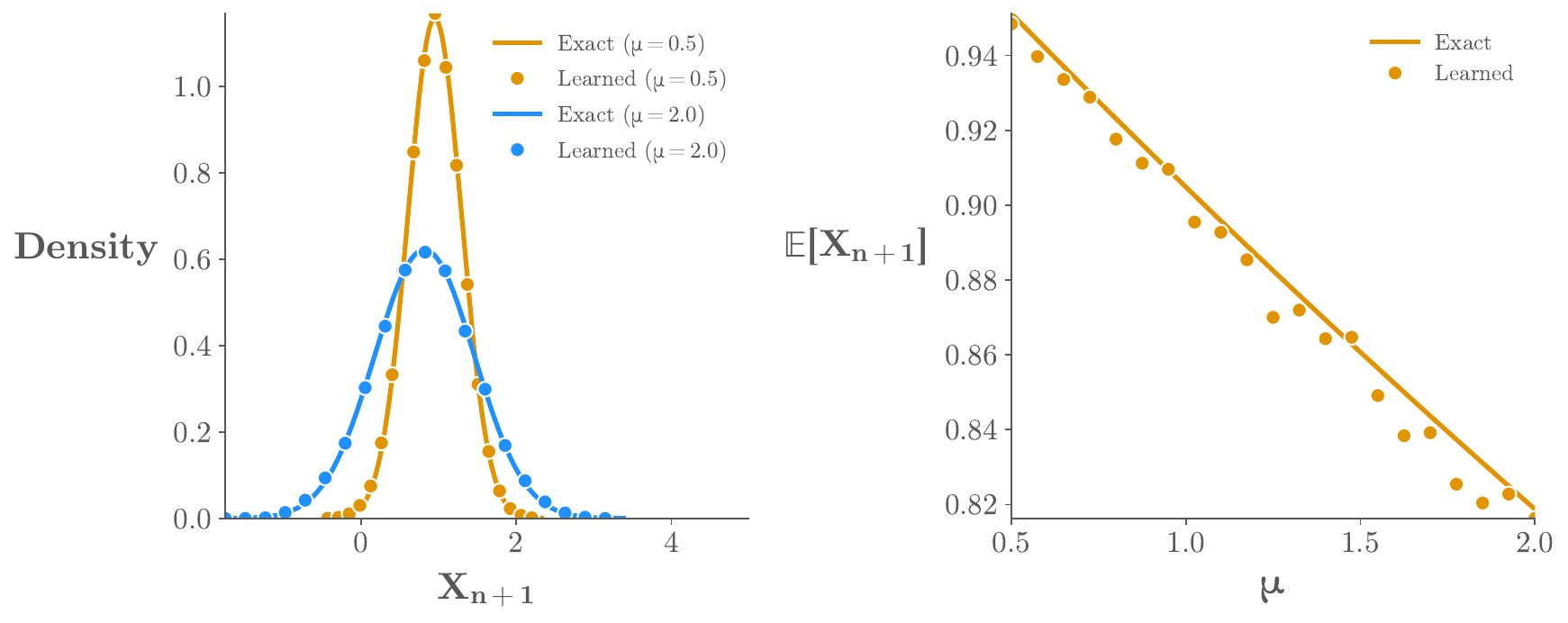}
\caption{Example 2: Conditional distribution $p(X_{n+1} \mid X_n = 1, \mu)$. Left: PDF comparison for $\mu = 0.5$ and $\mu = 2.0$; solid lines show exact Gaussian density \eqref{eq:ex2_exact_cond}, markers show learned estimates. Right: Conditional mean $\E[X_{n+1} \mid X_n = 1, \mu]$ vs.\ $\mu$; solid line is the exact formula $x e^{-\mu \Delta t}$, markers are learned estimates.}
\label{fig:ex2_conditional}
\end{figure}

Figure \ref{fig:ex2_conditional} shows the learned conditional distribution compared with the exact solution \eqref{eq:ex2_exact_cond}.
The left panel displays the PDF $p(X_{n+1} \mid X_n = 1, \mu)$ for $\mu = 0.5$ and $\mu = 2.0$.
The exact conditional means are $1 \cdot e^{-0.5 \cdot 0.1} \approx 0.951$ and $1 \cdot e^{-2 \cdot 0.1} \approx 0.819$, respectively.
The mean-reversion effect is clearly visible: larger $\mu$ leads to stronger attraction toward zero (smaller conditional mean), while the diffusion coefficient $\sqrt{1+\mu^2}$ also increases with $\mu$.
The learned distributions (markers) closely match the exact Gaussian distributions (solid lines).
The right panel shows the conditional mean as a function of $\mu$, demonstrating accurate capture of the exponential decay factor $e^{-\mu \Delta t}$.
Quantitatively, the maximum relative error in the conditional mean is $1.4\%$ (mean $0.6\%$), and the maximum relative error in the conditional variance is $2.8\%$ (mean $1.6\%$).

\begin{figure}[htbp]
\centering
\includegraphics[width=0.9\textwidth]{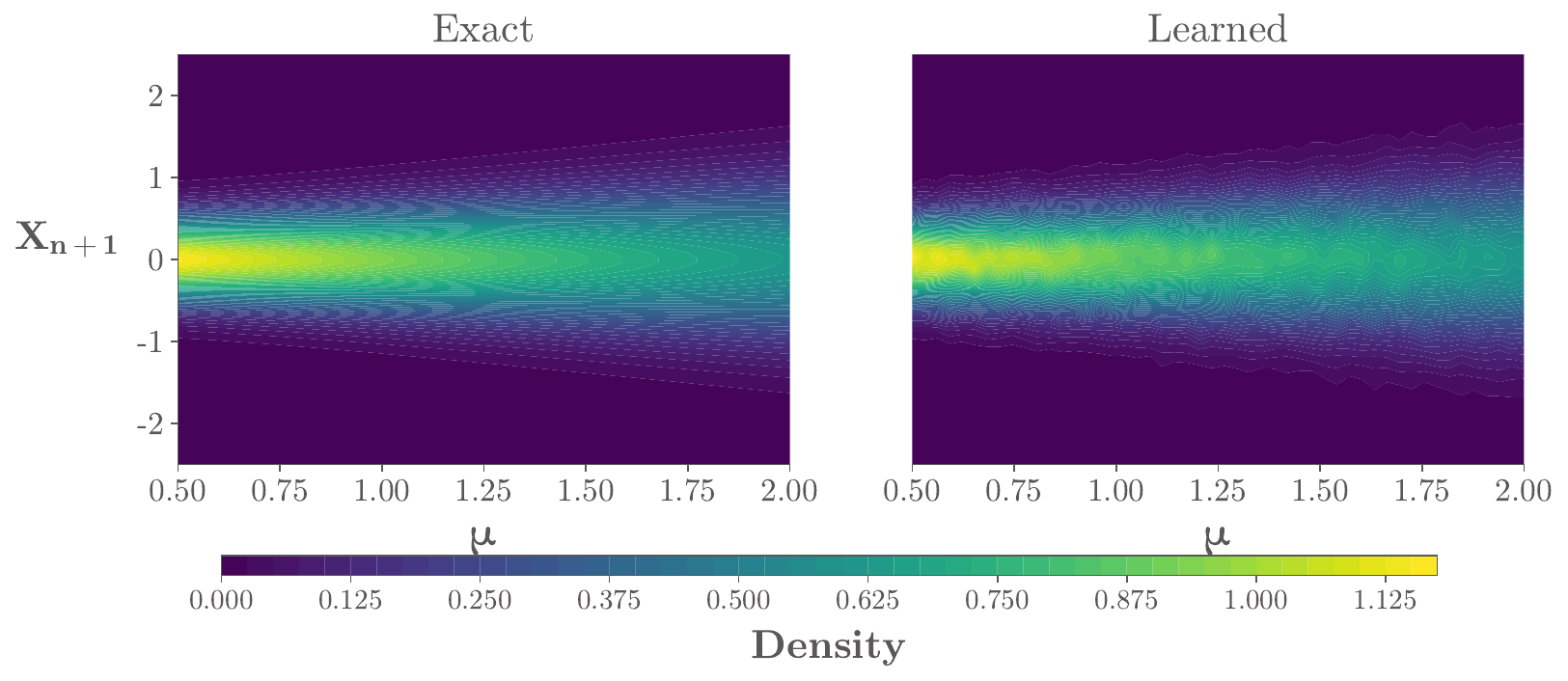}
\caption{Example 2: Heatmap of conditional distribution $p(X_{n+1} \mid X_n = 0, \mu)$. Left: Exact. Right: Learned. The one-step conditional variance increases monotonically with $\mu$. Constructed using $20{,}000$ samples per $\mu$ value.}
\label{fig:ex2_heatmap}
\end{figure}

Figure \ref{fig:ex2_heatmap} presents a heatmap of the conditional distribution $p(X_{n+1} \mid X_n = 0, \mu)$ starting from the equilibrium point.
The one-step conditional variance increases monotonically with $\mu$, as seen from the increasing spread of the distribution from left to right.
The learned model accurately captures this variance structure across the full parameter range.

\begin{figure}[htbp]
\centering
\includegraphics[width=0.9\textwidth]{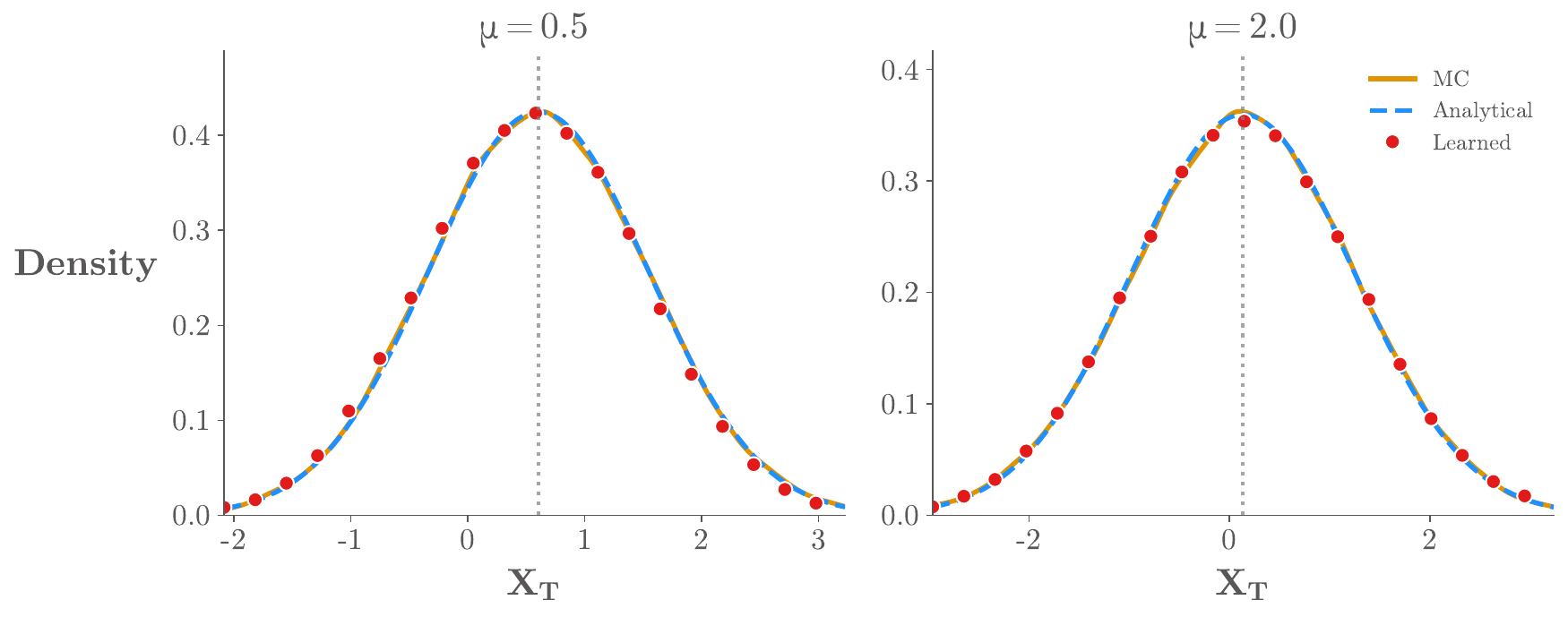}
\caption{Example 2: Terminal distribution $p(X_T)$ at $T = 1.0$ with Gaussian initial distribution $X_0 \sim \N(1.0, 0.25)$, obtained via 10 iterative applications of the learned one-step flow map. Solid lines: Monte Carlo ground truth ($50{,}000$ samples); dashed lines: analytical formula \eqref{eq:ex2_terminal}; markers: learned method ($50{,}000$ generated trajectories). Vertical dotted lines indicate the analytical mean $m_0 e^{-\mu T}$.}
\label{fig:ex2_terminal}
\end{figure}

Figure \ref{fig:ex2_terminal} shows the terminal distribution $p(X_T)$ at $T = 1.0$ starting from a non-equilibrium initial distribution $X_0 \sim \N(1.0, 0.25)$.
Multi-step trajectories are generated using $N_{\text{steps}} = 10$ applications of the learned flow map.
The figure compares three approaches: Monte Carlo ground truth (solid lines), analytical formula \eqref{eq:ex2_terminal} (dashed lines), and our learned method (markers).
The vertical dotted lines indicate the analytical mean $m_0 e^{-\mu T}$, which equals $0.607$ for $\mu = 0.5$ and $0.135$ for $\mu = 2.0$.
The mean-reversion toward zero is clearly visible, with stronger reversion for larger $\mu$.
The learned distributions closely match both references, with terminal mean errors below $0.06$ and standard deviation errors below $0.02$.

\begin{figure}[htbp]
\centering
\includegraphics[width=0.9\textwidth]{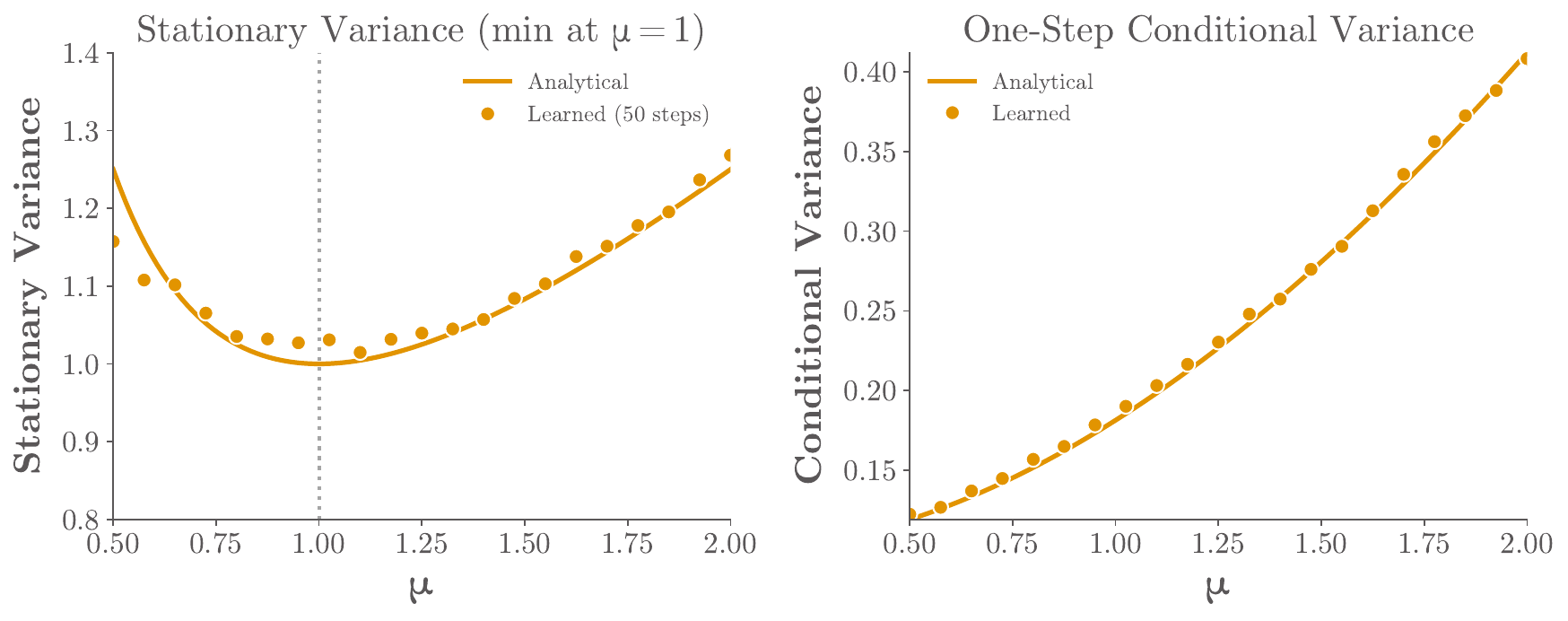}
\caption{Example 2: Variance analysis. Left: Stationary variance $(1+\mu^2)/(2\mu)$ showing non-monotonic behavior with minimum at $\mu=1$ (vertical dashed line). Learned variance computed from $50{,}000$ samples after 50 time steps. Right: One-step conditional variance from $X_n = 0$, comparing exact formula \eqref{eq:ex2_exact_cond} with learned estimates ($20{,}000$ samples per $\mu$ value).}
\label{fig:ex2_variance}
\end{figure}

Figure \ref{fig:ex2_variance} quantifies the non-monotonic variance behavior that makes this example particularly interesting.
The left panel shows the stationary variance $\sigma_\infty^2(\mu) = (1 + \mu^2)/(2\mu)$ as a function of $\mu$, which achieves its minimum value of $1$ at $\mu = 1$ (indicated by the vertical dotted line).
The learned model accurately captures this non-monotonic structure after 50 time steps of evolution ($50{,}000$ samples per $\mu$ value), with maximum relative error $7.4\%$ (mean $1.6\%$).
The right panel displays the one-step conditional variance, which increases monotonically with $\mu$.
The learned conditional variance has maximum relative error $3.4\%$ (mean $1.7\%$).
The excellent agreement between analytical and learned variances demonstrates that our method correctly captures both the mean and variance structure of the parameter-dependent dynamics.

\subsection{Example 3: Two-dimensional Ornstein-Uhlenbeck process with rotation}

As a final example demonstrating the method's capability in higher dimensions, we consider a two-dimensional Ornstein-Uhlenbeck process with rotation.
This example showcases genuinely coupled 2D dynamics where the parameter controls both the decay rate and the spiral behavior of trajectories.

\subsubsection{Problem formulation}

Consider the general SDE \eqref{eq:sde_general} with $d = m = 2$ and coefficients:
\begin{equation}\label{eq:ex3_coeffs}
a(x, \mu) = -Ax, \quad b(x, \mu) = \sigma I_2,
\end{equation}
where $\sigma = 0.5$ and the drift matrix is:
\begin{equation}\label{eq:ex3_matrix}
A = \begin{pmatrix} \mu & -\omega \\ \omega & \mu \end{pmatrix},
\end{equation}
with $\mu \in [0.5, 2.0]$ being the decay rate parameter and $\omega = 1.0$ the fixed rotation frequency.
This yields the two-dimensional SDE:
\begin{equation}\label{eq:ex3_sde}
\dd X_t = -A X_t \, \dd t + \sigma \, \dd W_t, \quad X_0 = x_0 \in \R^2,
\end{equation}
where $W_t$ is a standard 2D Brownian motion.
This matrix structure combines exponential decay (controlled by $\mu$) with rotation (controlled by $\omega$), resulting in spiral trajectories that converge to the origin.

\subsubsection{Exact conditional statistics}

The matrix exponential $e^{-At}$ can be computed analytically:
\begin{equation}\label{eq:ex3_matrix_exp}
e^{-At} = e^{-\mu t} \begin{pmatrix} \cos(\omega t) & \sin(\omega t) \\ -\sin(\omega t) & \cos(\omega t) \end{pmatrix}.
\end{equation}
This factorization shows that trajectories undergo exponential decay at rate $\mu$ while simultaneously rotating at angular velocity $\omega$.

The one-step conditional distribution is Gaussian:
\begin{align}
\E[X_{n+1} \mid X_n = x, \mu] &= e^{-A\Delta t} x, \label{eq:ex3_mean} \\
\text{Cov}[X_{n+1} \mid X_n = x, \mu] &= \frac{\sigma^2}{2\mu}\left(1 - e^{-2\mu\Delta t}\right) I_2. \label{eq:ex3_cov}
\end{align}
Due to the isotropic noise structure, the covariance matrix is diagonal with equal variances in both dimensions, despite the asymmetric rotation dynamics.

\subsubsection{Training procedure}

We generate training data using exact sampling from the analytical transition distribution with $\Delta t = 0.1$, simulation time $T = 2.0$, $N_\mu = 16$ parameter values uniformly spaced in $[0.5, 2.0]$, and $N_s = 3000$ trajectories per parameter value, yielding approximately $960{,}000$ training pairs in $\R^2$.

Following Algorithm \ref{alg:main}, we generate $M = 50{,}000$ labeled samples using $N = 2000$ nearest neighbors and $N_\tau = 500$ ODE time steps.
Due to the higher dimensionality, we use a larger bandwidth $\nu_x = 0.8$ and $\nu_\mu = 0.3$, with scaling factor $c_{\text{scale}} = 3.0$.
The neural network uses four hidden layers with $256$ neurons each to accommodate the increased complexity of 2D dynamics.

\subsubsection{Results}

\begin{figure}[htbp]
\centering
\includegraphics[width=0.95\textwidth]{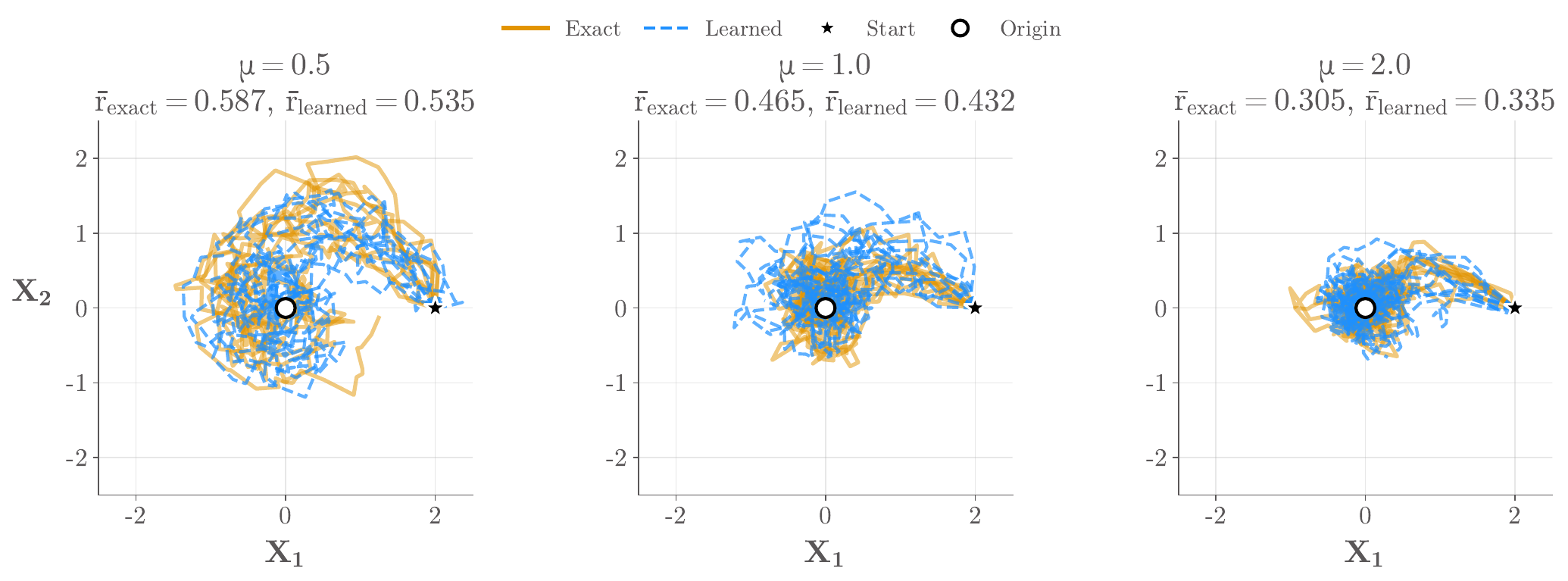}
\caption{Example 3: Sample trajectories for 2D OU with rotation starting from $(2, 0)$ over $T = 5$. Left to right: $\mu = 0.5$, $1.0$, $2.0$. Solid lines: exact SDE trajectories; dashed lines: learned model trajectories. The mean final radius $\bar{r}$ (averaged over 100 trajectories) measures convergence to the origin; larger $\mu$ leads to faster decay and smaller $\bar{r}$.}
\label{fig:ex3_trajectories}
\end{figure}

Figure \ref{fig:ex3_trajectories} shows sample trajectories for three decay rates: $\mu = 0.5$ (slow decay), $\mu = 1.0$ (moderate decay), and $\mu = 2.0$ (fast decay).
In all cases, trajectories spiral inward toward the origin due to the rotation term $\omega$.
The learned trajectories (dashed) closely follow the exact trajectories (solid), demonstrating that the model captures the coupled rotation-decay dynamics.
Each panel reports the mean final radius $\bar{r} = \frac{1}{N}\sum_{i=1}^{N} \|X_T^{(i)}\|$, computed as the sample average of the Euclidean distance from the origin at the final time $T = 5$ over $N = 100$ trajectories.
Larger $\mu$ leads to faster exponential decay, resulting in smaller $\bar{r}$.
The close agreement between $\bar{r}_{\text{exact}}$ (from exact simulations) and $\bar{r}_{\text{learned}}$ (from the learned model) confirms that the learned model correctly captures the parameter-dependent decay rate.

\begin{figure}[htbp]
\centering
\includegraphics[width=0.95\textwidth]{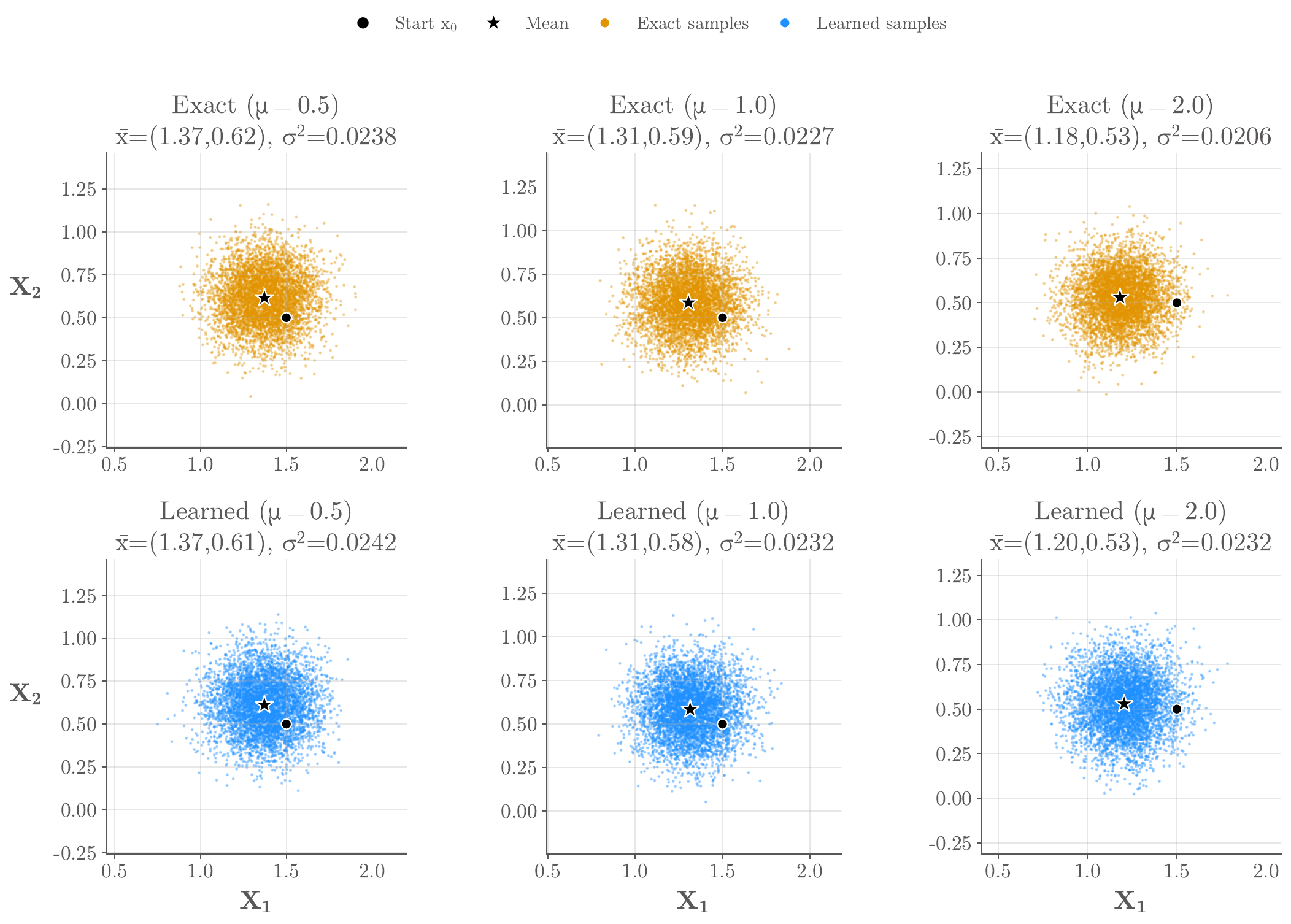}
\caption{Example 3: One-step conditional distribution scatter plots starting from $(1.5, 0.5)$. Top row: exact samples from analytical Gaussian \eqref{eq:ex3_mean}--\eqref{eq:ex3_cov}. Bottom row: learned model samples ($5{,}000$ samples each). Left to right: $\mu = 0.5$, $1.0$, $2.0$. Stars indicate the conditional mean; black dots indicate the starting position $x_0$.}
\label{fig:ex3_scatter}
\end{figure}

Figure \ref{fig:ex3_scatter} shows the one-step conditional distribution starting from $x_0 = (1.5, 0.5)$ for three $\mu$ values.
The top row shows exact samples from the analytical Gaussian distribution, and the bottom row shows learned samples.
The learned model accurately captures both the location (mean) and spread (covariance) of the distribution, with the characteristic rotation visible in how the mean shifts relative to the starting point.
From the conditional covariance formula \eqref{eq:ex3_cov}, the one-step variance is approximately $\sigma^2 \Delta t$ for small $\Delta t$, but decreases with larger $\mu$ due to the factor $1/(2\mu)$.
Quantitatively, the mean position error is below $0.024$ and the relative variance error is below $3\%$ for $\mu \leq 1$.
For $\mu = 2$, the relative variance error increases to $12.6\%$; however, this larger relative error reflects the smaller absolute variance at high $\mu$ (the conditional variance at $\mu = 2$ is approximately half that at $\mu = 0.5$), making relative errors more sensitive to small absolute differences.

\begin{figure}[htbp]
\centering
\includegraphics[width=0.9\textwidth]{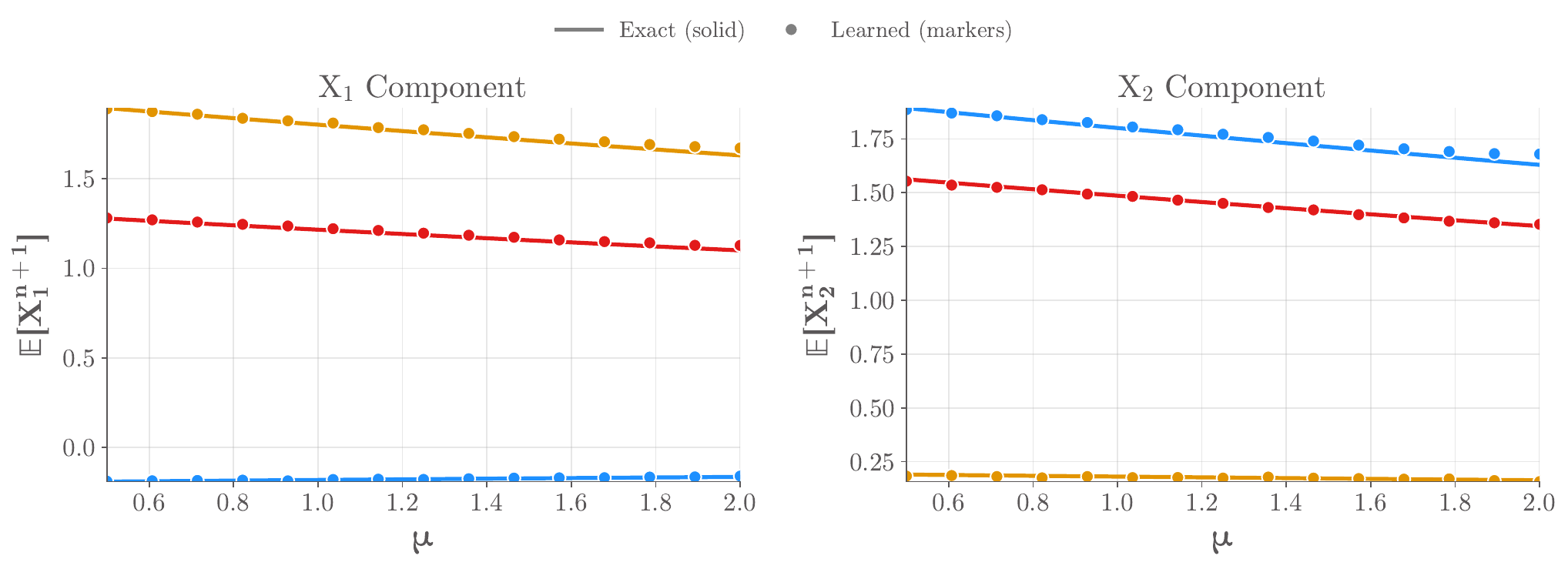}
\caption{Example 3: Conditional mean comparison for 2D OU with rotation. Left: $X_1$ component. Right: $X_2$ component. Solid lines: exact formula \eqref{eq:ex3_mean}. Markers: learned estimates ($5{,}000$ samples per point). Three starting positions are shown: $(2, 0)$, $(0, 2)$, and $(1.5, 1.5)$.}
\label{fig:ex3_mean}
\end{figure}

Figure \ref{fig:ex3_mean} compares the learned conditional mean with the exact formula \eqref{eq:ex3_mean} for three starting positions.
The left panel shows the $X_1$ component and the right panel shows the $X_2$ component.
The learned means (markers) closely track the exact curves (solid lines) across the full parameter range, demonstrating accurate learning of the rotation dynamics.
The maximum absolute error in the conditional mean is below $0.05$ for all starting positions and parameter values.

\begin{figure}[htbp]
\centering
\includegraphics[width=0.9\textwidth]{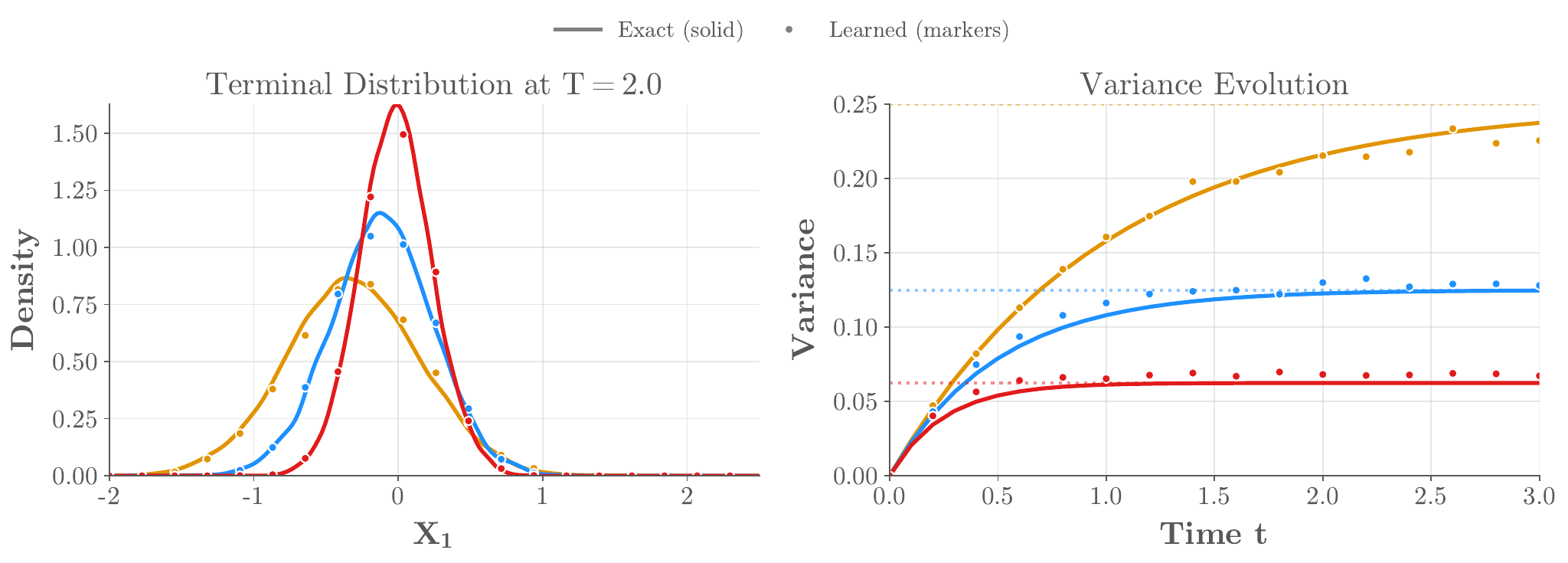}
\caption{Example 3: Multi-step distribution and variance analysis. Left: Marginal distribution of $X_1$ at $T = 2$ for $\mu = 0.5$, $1.0$, $2.0$. Right: Variance evolution over time starting from deterministic initial condition $x_0 = (2, 0)$. Variance grows from zero toward the stationary variance $\sigma^2/(2\mu)$, which depends strongly on $\mu$. Solid lines: exact analytical formula; markers: learned model.}
\label{fig:ex3_multistep}
\end{figure}

Figure \ref{fig:ex3_multistep} shows the multi-step behavior.
The left panel displays the marginal distribution of $X_1$ at time $T = 2$ for three $\mu$ values.
The learned distributions (markers) closely match the exact Monte Carlo distributions (solid lines).
The right panel shows the variance evolution over time starting from a deterministic initial condition.
Variance grows from zero toward the stationary value $\sigma^2/(2\mu)$, which is $0.25$ for $\mu=0.5$, $0.125$ for $\mu=1.0$, and $0.0625$ for $\mu=2.0$.
The learned variance trajectories (markers) accurately track the exact analytical curves (solid lines), demonstrating that the model correctly captures the parameter-dependent diffusion dynamics.
For the terminal distribution, the mean error is below $0.04$ and the relative variance error is below $8\%$.

This example demonstrates that the proposed method scales effectively to multi-dimensional SDEs with coupled dynamics.
The learned model captures both the rotational structure (from the off-diagonal terms in $A$) and the parameter-dependent decay rate, producing accurate samples from the joint 2D conditional distribution.

\section{Conclusion}\label{sec:conclude}

This work presents a training-free conditional diffusion framework for learning stochastic flow maps of parameter-dependent SDEs of the form $\dd X_t = a(X_t, \mu)\,\dd t + b(X_t, \mu)\,\dd W_t$.
The key innovation lies in extending the training-free score estimation approach of \cite{liu2024training, yang2025generative} to handle parameter dependence, where the conditional score function is approximated using Gaussian kernel-weighted combinations of empirical displacements.
This enables efficient generation of labeled training data for neural network regression without solving a computationally expensive score-matching optimization problem, which is particularly advantageous when exploring large parameter spaces.

Our methodology offers several distinct advantages for learning parameter-dependent stochastic dynamics.
The training-free score estimation eliminates the computational overhead associated with neural network-based score function learning, replacing it with a closed-form Monte Carlo estimator that directly uses trajectory data.
By incorporating parameter dependence through kernel weighting in both spatial and parameter dimensions, the framework learns a unified model that captures how the conditional distribution varies with the parameter $\mu$, rather than requiring separate models for each parameter value.
The two-stage approach---first generating labeled data via probability flow ODE, then training a simple neural network via supervised regression---provides training stability and architectural flexibility that would be difficult to achieve with end-to-end generative model training.

We demonstrate the framework's effectiveness through comprehensive validation across three benchmark problems of increasing complexity.
The parameter-dependent Brownian motion (Example 1) validates the basic methodology, with the learned model achieving mean absolute errors below $0.006$ for the conditional mean and relative errors below $3.5\%$ for the conditional variance across the entire parameter range.
The Ornstein-Uhlenbeck process with parameter-dependent diffusion (Example 2) tests the method's ability to capture non-trivial parameter dependence, where the stationary variance exhibits non-monotonic behavior as a function of the mean-reversion rate; the learned model reproduces this behavior with relative variance errors below $3.5\%$.
The two-dimensional OU process with rotation (Example 3) extends the validation to multi-dimensional coupled dynamics, where the learned model captures both the rotational structure from off-diagonal drift terms and the parameter-dependent decay rate, achieving conditional mean errors below $1.5\%$.
In all cases, multi-step trajectory generation via composition of single-step flow maps produces accurate long-time statistics, including terminal distributions and time-dependent variance evolution, confirming the method's ability to generalize beyond single-step predictions.

Future work will focus on extending the framework to higher-dimensional systems, which will require developing more efficient strategies for kernel density estimation and neighbor search.
Additionally, we will investigate adaptive strategies for selecting the bandwidth parameters $\nu_x$ and $\nu_\mu$; connections to optimal bandwidth selection in kernel density estimation may provide theoretical guidance for automatic tuning.
Theoretical analysis of the approximation error introduced by the training-free score estimation would establish rigorous bounds on the learned flow map accuracy.
Finally, the learned stochastic flow maps could serve as surrogate models in data assimilation frameworks, where accurate parameter-dependent dynamics are essential for state estimation in systems with uncertain parameters.

\section*{Acknowledgments}
This material is based upon work supported in part by the U.S.~Department of Energy, Office of Science, Offices of Advanced Scientific Computing Research and Fusion Energy Science, and by the Laboratory Directed Research and Development program at the Oak Ridge National Laboratory, which is operated by UT-Battelle, LLC, for the U.S.~Department of Energy under Contract DE-AC05-00OR22725. S.\ He acknowledges support from the University of Tennessee, Knoxville AI seed grant.


\bibliographystyle{elsarticle-num}
\bibliography{bib/refs,bib/claude}


\appendix

\section{Forward and reverse diffusion processes}\label{app:diffusion}

This appendix provides the theoretical foundations for the parameter-dependent conditional diffusion model used in Section~\ref{sec:Method}.

\subsection{Forward diffusion process}

For a fixed pair $(x_n, \mu)$, let $\Delta X_{n+1}^{x_n,\mu} = X_{n+1}^{x_n,\mu} - x_n$ denote the displacement, where $X_{n+1}^{x_n,\mu} \sim p(\cdot \mid x_n, \mu)$ follows the conditional distribution.
Following \cite{liu2024training}, we define a forward diffusion process on the displacement in an artificial domain $\tau \in [0, 1]$:
\begin{equation}\label{eq:forward_sde}
\dd Z_\tau^{x_n,\mu} = f(\tau) Z_\tau^{x_n,\mu} \, \dd \tau + g(\tau) \, \dd W_\tau, \quad Z_0^{x_n,\mu} = \Delta X_{n+1}^{x_n,\mu},
\end{equation}
where $W_\tau$ is a standard Brownian motion (independent of the original SDE), and $f(\tau)$ and $g(\tau)$ are the drift and diffusion coefficients of the forward diffusion process.

We choose the coefficients such that the forward process has an analytically tractable transition density.
Specifically, we parameterize the solution of \eqref{eq:forward_sde} as
\begin{equation}\label{eq:zt_form}
Z_\tau^{x_n,\mu} = \alpha_\tau Z_0^{x_n,\mu} + \beta_\tau \epsilon, \quad \epsilon \sim \N(0, I_d),
\end{equation}
where $\alpha_\tau$ and $\beta_\tau$ are deterministic scaling functions satisfying $\alpha_0 = 1$, $\beta_0 = 0$.
For this parameterization to be consistent with \eqref{eq:forward_sde}, the coefficients must satisfy:
\begin{equation}\label{eq:diffusion_coeffs}
f(\tau) = \frac{\dd \log \alpha_\tau}{\dd \tau}, \quad g^2(\tau) = \frac{\dd \beta_\tau^2}{\dd \tau} - 2 \frac{\dd \log \alpha_\tau}{\dd \tau} \beta_\tau^2.
\end{equation}
These relations ensure that the marginal distribution of $Z_\tau$ follows the prescribed scaling.

In this work, we adopt the variance-preserving (VP) schedule \cite{ho2020denoising}:
\begin{equation}\label{eq:scaling_functions}
\alpha_\tau = 1 - \tau, \quad \beta_\tau^2 = \tau, \quad \tau \in [0, 1].
\end{equation}
With this choice, we have $\alpha_1 = 0$ and $\beta_1 = 1$, so that $Z_1^{x_n,\mu} \sim \N(0, I_d)$ regardless of the initial displacement distribution $p_\Delta(\cdot \mid x_n, \mu)$.

From \eqref{eq:zt_form}, the conditional distribution of $Z_\tau^{x_n,\mu}$ given $Z_0^{x_n,\mu} = \Delta x$ is Gaussian:
\begin{equation}\label{eq:conditional_gaussian}
Q(Z_\tau^{x_n,\mu} \mid Z_0^{x_n,\mu} = \Delta x) = \N(\alpha_\tau \Delta x, \beta_\tau^2 I_d).
\end{equation}
This result follows directly from the linearity of \eqref{eq:zt_form}: given $Z_0^{x_n,\mu} = \Delta x$, we have $Z_\tau^{x_n,\mu} = \alpha_\tau \Delta x + \beta_\tau \epsilon$ where $\epsilon \sim \N(0, I_d)$, yielding a Gaussian with mean $\alpha_\tau \Delta x$ and covariance $\beta_\tau^2 I_d$.

\subsection{Reverse-time SDE and ODE}

The key result from score-based diffusion theory is that the forward diffusion process \eqref{eq:forward_sde} has a corresponding reverse-time process.
Running the reverse process from $\tau = 1$ to $\tau = 0$ transforms samples from $\N(0, I_d)$ back to samples from the target displacement distribution $p_\Delta(\cdot \mid x_n, \mu)$.

Following~\cite{anderson1982reverse}, the reverse-time SDE is
\begin{equation}\label{eq:reverse_sde}
\dd Z_\tau^{x_n,\mu} = \left[ f(\tau) Z_\tau^{x_n,\mu} - g^2(\tau) S(Z_\tau^{x_n,\mu}, \tau; x_n, \mu) \right] \dd \tau + g(\tau) \, \dd \overleftarrow{W}_\tau,
\end{equation}
where $\overleftarrow{W}_\tau$ is a backward Brownian motion, and $S(z, \tau; x_n, \mu)$ is the parameter-dependent conditional score function defined in \eqref{eq:score_def}.
Note that the score function depends on four quantities: the diffusion state $z$, the diffusion time $\tau$, the initial state $x_n$, and the parameter $\mu$; the dependence on $\mu$ is the key extension from the standard conditional diffusion model in \cite{liu2024training}.

An important observation from \cite{song2020score} is that the reverse-time SDE \eqref{eq:reverse_sde} shares the same marginal distributions $Q(Z_\tau^{x_n,\mu} \mid x_n, \mu)$ with the deterministic probability flow ODE \eqref{eq:reverse_ode} presented in Section~\ref{sec:Method}.
The ODE defines a deterministic mapping from $Z_1 \sim \N(0, I_d)$ to $Z_0 \sim p_\Delta(\cdot \mid x_n, \mu)$, which is essential for generating labeled training data with a one-to-one correspondence between latent variables and displacement outputs.

\section{Score function derivation}\label{app:score}

This appendix provides the complete derivation of the closed-form expression for the parameter-dependent conditional score function used in Section~\ref{sec:Method}.

The marginal density of $Z_\tau^{x_n,\mu}$ conditioned on $(x_n, \mu)$ can be obtained by integrating over all possible initial displacement values:
\begin{equation}\label{eq:marginal_density}
Q(Z_\tau^{x_n,\mu} = z \mid x_n, \mu) = \int_{\R^d} Q(Z_\tau^{x_n,\mu} = z \mid Z_0^{x_n,\mu} = \Delta x) \, p_\Delta(\Delta x \mid x_n, \mu) \, \dd (\Delta x).
\end{equation}
Substituting the Gaussian transition density \eqref{eq:conditional_gaussian}, we have
\begin{equation}\label{eq:marginal_explicit}
Q(z \mid x_n, \mu) = \int_{\R^d} \frac{1}{(2\pi \beta_\tau^2)^{d/2}} e^{-\frac{\|z - \alpha_\tau \Delta x\|^2}{2\beta_\tau^2}} p_\Delta(\Delta x \mid x_n, \mu) \, \dd (\Delta x).
\end{equation}
This integral representation expresses the marginal density as a mixture of Gaussians weighted by the displacement distribution $p_\Delta(\Delta x \mid x_n, \mu)$.

To derive the score function, we apply the log-derivative:
\begin{equation}\label{eq:score_derivation_1}
S(z, \tau; x_n, \mu) = \nabla_z \log Q(z \mid x_n, \mu) = \frac{\nabla_z Q(z \mid x_n, \mu)}{Q(z \mid x_n, \mu)}.
\end{equation}
Taking the gradient of \eqref{eq:marginal_explicit} with respect to $z$:
\begin{equation}\label{eq:gradient_Q}
\nabla_z Q(z \mid x_n, \mu) = \int_{\R^d} \left( -\frac{z - \alpha_\tau \Delta x}{\beta_\tau^2} \right) Q(z \mid \Delta x) \, p_\Delta(\Delta x \mid x_n, \mu) \, \dd (\Delta x),
\end{equation}
where $Q(z \mid \Delta x) \coloneqq Q(Z_\tau^{x_n,\mu} = z \mid Z_0^{x_n,\mu} = \Delta x) = \N(z; \alpha_\tau \Delta x, \beta_\tau^2 I_d)$.

Substituting \eqref{eq:marginal_explicit} and \eqref{eq:gradient_Q} into \eqref{eq:score_derivation_1}, we obtain the exact score function:
\begin{equation}\label{eq:score_exact}
S(z, \tau; x_n, \mu) = \frac{\int_{\R^d} \left( -\frac{z - \alpha_\tau \Delta x}{\beta_\tau^2} \right) Q(z \mid \Delta x) \, p_\Delta(\Delta x \mid x_n, \mu) \, \dd (\Delta x)}{\int_{\R^d} Q(z \mid \Delta x') \, p_\Delta(\Delta x' \mid x_n, \mu) \, \dd (\Delta x')}.
\end{equation}
This expression shows that the score is a ratio of weighted integrals over the displacement distribution.

This can be rewritten in a more interpretable form by defining the weight function:
\begin{equation}\label{eq:weight_function}
w_\tau(z, \Delta x; x_n, \mu) \coloneqq \frac{Q(z \mid \Delta x)}{\int_{\R^d} Q(z \mid \Delta x') \, p_\Delta(\Delta x' \mid x_n, \mu) \, \dd (\Delta x')},
\end{equation}
which satisfies the normalization $\int w_\tau \, p_\Delta(\Delta x \mid x_n, \mu) \, \dd (\Delta x) = 1$.

With this notation, the score function becomes equation \eqref{eq:score_weighted} in the main text.
This representation reveals that the score function is a weighted average of ``local scores'' $-(z - \alpha_\tau \Delta x)/\beta_\tau^2$ over all possible displacement values, where the weights favor displacements that are more likely to have produced the current diffusion state $z$.

\end{document}